\documentclass[lettersize,journal]{IEEEtran}
\usepackage{amsmath,amsfonts}
\usepackage{algorithmic}
\usepackage{algorithm}
\usepackage{array}
\usepackage{subcaption}
\usepackage{textcomp}
\usepackage{stfloats}
\usepackage{url}
\usepackage{verbatim}
\usepackage{graphicx}
\usepackage{cite}
\usepackage{enumitem}
\usepackage{booktabs}
\usepackage{diagbox}
\usepackage{color}

\usepackage{changes}
\usepackage{hyperref}
\usepackage{tabularx}
\usepackage{colortbl}
\usepackage{array}
\usepackage{pdfpages} % put this in your preamble

\usepackage{multirow,graphicx,xcolor}

\begin{document}

\title{Fr\'{e}chet Power-Scenario Distance: A Metric for Evaluating Synthetic Data Across Multiple Time-Scales in Smart Grids}

\author{Yuting Cai,~\IEEEmembership{Student Member,~IEEE,} Shaohuai Liu,~\IEEEmembership{Student Member,~IEEE,} Chao Tian,~\IEEEmembership{Senior Member,~IEEE}, Le Xie,~\IEEEmembership{Fellow,~IEEE}
        % <-this % stops a space
\thanks{Yuting Cai, Shaohuai Liu and Chao Tian are with Department of Electrical and Computer Engineering, Texas A\&M University, College Station, TX, 77843 USA (e-mail: cyuting@tamu.edu;
liushaohuai5@tamu.edu; chao.tian@tamu.edu). Le Xie is with the John A. Paulson School of Engineering and Applied Sciences, Harvard University, Boston, MA 02134 USA (e-mail:
xie@seas.harvard.edu)}}% <-this % stops a space



\maketitle

\begin{abstract}
Generative artificial intelligence (AI) models in smart grids have advanced significantly in recent years due to their ability to generate large amounts of synthetic data, which would otherwise be difficult to obtain in the real world due to confidentiality constraints. A key challenge in utilizing such synthetic data is how to assess the data quality produced from such generative models.
Traditional metrics such as sample-wise Euclidean distance and distributional distances applied directly to raw generated data inadequately reflect higher-order temporal dependencies and cross-temporal relationships between real and synthetic series, and thus struggle to discriminate generative quality.
% Traditional \emph{Euclidean distance-based} metrics only reflect pair-wise relations between two individual samples, even the distribution-wise metric, such as Energy Score, can not capture the high-level temporal or cross-temporal similarity between real and synthetic data, and could fail in evaluating quality differences between groups of synthetic datasets.
% \added{Traditional Euclidean distance-based metrics only reflect pair-wise relations between two samples and could fail in evaluating quality differences between groups of generations. }
In this work, we propose a novel metric based on the \emph{Fr\'{e}chet Distance (FD)} estimated between two datasets in a learned feature space. The proposed method assesses synthetic data quality via distributional comparisons in a feature space derived from a model tailored to the smart grid domain.
% \added{The proposed method evaluates the generation quality in a distributional perspective.} 
Empirical results demonstrate the superiority of the proposed metric across downstream tasks and generative models, enhancing the reliability of data-driven decision-making in smart grid operations.

% We verify empirically the efficacy of the proposed metric using extensive experiments across multiple timescales in power systems.
\end{abstract}


\section{Introduction}
\IEEEPARstart{G}{enerative} models in the electric energy sector have been an active field of research in the past few years, thanks to their potential to create realistic and diverse scenarios for system planning, reliability assessment, and renewable energy integration—ultimately enhancing grid resilience and operational efficiency. These models, such as Generative Adversarial Networks (GANs), allow researchers to access much larger sets of synthetic data across multiple time scales that would otherwise be unavailable due to confidentiality constraints~\cite{yuan2022conditional}. In contrast to traditional methods that involve creating synthetic power networks and subsequently using commercial-grade simulation software to generate electrical measurement variables~\cite{7725528}, these generative approaches leverage a data-driven methodology. They produce large volumes of synthetic electrical data by utilizing historical electrical measurements in a computationally efficient manner.

While these generative models hold great potential for a wide range of power system applications, a central question remains largely unexplored: What would be a fair and robust metric to evaluate the quality of various generative techniques in creating synthetic power system data across multiple time resolutions? Conventionally, evaluation methods are based on Euclidean metrics between pairs of samples (signals), such as Mean Squared Error (MSE) and Mean Absolute Percentage Error (MAPE)~\cite{bendaoud2021comparing,claeys2024capturing,liang2019sequence,liu2023generative}. However, these methods are susceptible to the selection of samples, leading to inconsistent evaluations. Moreover, the sample-by-sample comparison approach poses challenges when evaluating datasets produced by generative models, where the distributional properties of the datasets are more critical.

% As an alternative, distribution-based comparison methods, such as proper score rules (Continuous Ranked Probability Score (CRPS) or Energy Score (ES)) and Distribution–distribution distances (KL-divergence, MMD, Wasserstein) which can capture the broader distributional characteristics of generative models, have been explored in previous work \cite{razghandi2022variational}.
As an alternative, distribution-based approaches—covering proper scoring rules such as the Continuous Ranked Probability Score (CRPS) \cite{lemos2021probabilistic} and Energy Score (ES)\cite{pinson2012evaluating}, and distribution-to-distribution distances such as Kullback–Leibler (KL) divergence \cite{wang2020generating}, Maximum Mean Discrepancy (MMD)\cite{wang2024customized}, and Wasserstein distance \cite{razghandi2023smart}—have been explored in prior work to capture the broader distributional characteristics of synthetic datasets.
% generative models. 
However, when computed directly on raw data, distribution-level metrics tend to underweight higher-order structures such as temporal dependencies, which are crucial in smart grid time-series. Consequently, they can fail to discriminate meaningful changes in generative quality that primarily manifest in the dependence structure \cite{bjerregaard2021introduction}.
% Distance (FD), which is a metric for comparing two Gaussian distributions, 
% is used in some previous work such as Razghandi et al.\cite{razghandi2022variational}. 
% However, applying FD directly at the data level may not effectively capture the similarity between datasets.

Other approaches have opted for task-specific metrics that involve statistical analysis of task-based parameters. For instance, the number of zeros is employed as a task-specific indicator to assess the quality of voltage sag generation in \cite{liu2021methodologies}. Similarly, the indicators used to detect anomalies in electricity consumption include residential stimulus allocation, residential tariff allocation, and SME (Small and Medium-sized Enterprise) allocation in \cite{oprea2021anomaly}. However, these approaches introduce inconsistencies, as the evaluation criteria vary across tasks. Such variability makes it difficult to form a unified view of overall synthetic data quality or to establish standardized benchmarks, potentially leading to biased assessments.

Building on insights from computer vision research, where generative model evaluation has been extensively explored, we investigate methods that could establish more reliable evaluation standards. One such approach is the Inception Score (IS)~\cite{salimans2016improved}, which was initially adopted to assess the quality and diversity of generated images using a pretrained Inception network. However, IS faced criticism for its inability to detect the mode collapse of the generative model, where the model produces a limited variety of outputs or concentrates on only a few modes within the distribution, resulting in potential failures across various scenarios~\cite{9312049}. The Fr\'{e}chet Inception Distance (FID) was proposed to overcome these limitations. FID evaluates distributional similarity by taking the Fr\'{e}chet (squared 2-Wasserstein) distance between the Gaussian embeddings of real and generated images in feature spaces extracted by a pretrained Inception network. \cite{heusel2017gans}
% , namely Inception v3. This metric considers both the mean and covariance of the feature space of real and generated data distributions, making it more sensitive to variations in both quality and diversity while closely aligning with human perception \cite{heusel2017gans}. \cite{binkowski2018demystifying} suggests using Kernel Inception Distance (KID), which implements Maximum Mean Discrepancy (MMD) with kernel functions applied to the feature space of the Inception model can further improve the metric performance compared to FID. Unlike FID, which can suffer from estimator bias. KID provides an unbiased estimate, ensuring stable evaluation, particularly for small sample datasets. However, this approach is highly dependent on the choice of kernel functions, with different kernels potentially leading to inconsistent results.  
% \textcolor{red}{Need to mention KID as well. }

% Inspired by the success of the FID metric, in this paper we aim to develop a robust metric that could work in the context of multi-time-scale generative data sets in smart grids. The challenge that arises in adapting FID for this context are as follows: 

Inspired by the success of the FID metric, we argue that computing a distributional metric (e.g., Fr\'{e}chet distance) in a learned feature space yields a more comprehensive and fair evaluation of smart grid synthetic datasets.
The central challenge is thus obtaining a valid feature extractor that provides features aligned with the smart grid characteristics. Inception-v3, the encoder used by FID, is optimized for images rather than time-series. Thus, when applied to smart grid data, it underrepresents temporal dynamics. Moreover, its fixed \(299\times299\) input requirement is incompatible with the inherently multi-resolution, multi-duration nature of power-system signals. A natural alternative is to consider time-series encoders explicitly designed to capture temporal dynamics and to be flexible with respect to input length.

% As the power signals mostly appear in time-series, we can not directly adopt the feature extractor Inception-v3 used in FID because Inception-v3 is optimized for image features, not time-series structure, applying this to power-system data underrepresents the temporal and cross-site patterns central to grid behavior. Moreover, it operates on a fixed input resolution of \(299\times299\) pixels, which is incompatible with the inherently multi-resolution and multi-duration nature of power-system data. 

Although prior work has produced generic time-series encoders (e.g., TS2Vec \cite{yue2022ts2vec}, TimeMixer \cite{wang2024timemixer}), they are not plug-and-play for smart grids. While these encoders can handle variable sequence lengths—typically by padding or truncating to a fixed window—their learned representations often emphasize broad statistical or sequential regularities and miss physics-driven structure. In contrast to generic time-series, smart grid signals exhibit distinctive structure: strong diurnal and seasonal periodicities, hard physical constraints (e.g., ramp limits, nonnegativity, no solar at night), and a multi-resolution character. 
Capturing these attributes is crucial for downstream tasks such as anomaly detection, disturbance classification, and evaluation of generative scenarios, yet generic pretrained encoders frequently fail to preserve them.

% We emphasize that it cannot be directly applied to power-system data. Instead, it is necessary to design a new metric that builds on the underlying principles of FID but is specifically tailored to the unique characteristics of multi–time-scale generative datasets in smart grids. 

To address this gap, we propose a power-scenario-aware feature extractor that explicitly encodes the temporal and spectral signatures of grid data. Leveraging a multi-resolution hierarchical architecture and trained on large-scale power datasets, the extractor uses a classification head to emphasize local dynamics/pattern regimes and a regression head to capture broader statistical trends. Together, these objectives yield representations that faithfully capture the physical underpinnings of power-system data. Computing a distributional distance (e.g., Fr\'{e}chet) in this feature space then provides a more insightful, effective, and robust evaluation of power-domain generative models.

The key contributions of this paper are as follows: 
\begin{enumerate}[label=\arabic*)]
\item \textbf{A hierarchical multi-resolution feature extraction network:} We present a novel network design where multiple feature extractors are trained to handle data at different time resolutions and are connected in a hierarchical structure. This framework accommodates multi-resolution inputs, propagating features learned at shorter timescales to longer horizons. Each extractor is trained on smart grid datasets using joint classification and regression tasks, encouraging representations that capture source-specific dynamics as well as general statistical structure. Collectively, this improves the model's capacity to represent the complex temporal behavior inherent in smart grids.

\item \textbf{A feature-based, temporally aware distributional metric for generative models in smart grids:} 
Unlike traditional metrics that operate directly on raw data, we propose a metric that evaluates in a learned feature space: specifically, we compute the Fr\'{e}chet distance between extracted feature embeddings of real and generated data to quantify generative quality. Extensive experiments show that this approach robustly and consistently handles multi-timescale data and provides a unified standard across diverse smart grid generative tasks. We refer to this metric as the Fr\'{e}chet Power-Scenario Distance (FPD).

% It can effectively handle multi-timescale data and provides a unified standard for a diverse range of power system tasks.
% To our best knowledge, this is first-of-its-kind introduction of the Fr\'{e}chet Power-Scenario Distance (FPD) in the smart grids context.  FPD is specifically developed and trained to assess the performance of generative models in the power system domain. It can effectively handle multi-timescale data and provides a unified standard for a diverse range of power system tasks.
\end{enumerate}

The rest of the paper is organized as follows: Section \ref{section:design} details the design of the proposed evaluation metrics; Section \ref{section:training} introduces model training details; Section \ref{section:validation} demonstrates the effectiveness of FPD across various disturbances and its advantages over traditional metrics; and Section \ref{section:conclusion} concludes the paper.

\section{Design of Evaluation Metrics}
\label{section:design}
\subsection{Overview of the feature extraction model}
We employ neural networks to extract high-level features capable of capturing complex spatial and temporal relations.
% A similar approach is implemented in the Fr\'{e}chet Inception Distance (FID), where the feature extractor is a pre-trained neural network known as Inception V3. 
% The semantic features encoded by Inception V3 are sufficient to classify images accurately, making these high-level feature representations strong indicators of the descriptive power of the model.
Currently, a general feature extractor for this task does not exist in the smart grid domain. Implementing such a general model is challenging due to the diverse time resolutions and durations inherent in smart grid data. To address this challenge, we build a unified feature extraction framework that accommodates such diversity. 
As shown in Fig.~\ref{fig: Feature extraction model overview}, our proposed framework is a multi-resolution, multi-duration model, where a series of feature extractors, each trained for a specific time resolution, are interconnected hierarchically. Every feature extractor during training is paired with a classification head that emphasizes local dynamics and a regression head that captures broader statistical trends, together ensuring that the extracted features faithfully represent the characteristics of smart grid data. This design allows multiple extractors to be combined for different data durations, with output features from each higher-resolution extractor feeding into the input of the subsequent lower-resolution extractor. This integration of high-resolution features into lower-resolution modules enables effective use of fine-grained information while avoiding substantial increases in dimensional complexity. At the same time, it ensures that the output features at each level remain closely aligned with the characteristic patterns of smart grid datasets.

\subsection{Hierarchical feature extraction model}

Considering the different timescales
% \replaced{timescales }{physical characteristics} 
underlying the steady-state and transient-state data in smart grids, we propose two types of feature extraction models
% \replaced{feature extractors }{models}, 
respectively. Due to the different sampling intervals, there are multiple resolutions and durations in steady-state data. Therefore, we adopt a multi-layer hierarchical architecture. For the transient-state situations, the sampling intervals are much simpler, typically a selection of integer multiples of the rated frequency. Hence, we design a single-layer model for the transient-state case.
% \added{Due to the different sampling intervals, there are multiple resolutions and durations in steady-state data. Therefore, we adopt a multi-layer recursive architecture. For the transient-state situations, the sample intervals are much more straightforward, typically a selection of integer multiples of the rated frequency. Hence, we design a single-layer model for the transient-state case.}

% $ \deleted{and the durations $d \in \{\text{1 hour}, \text{1 day}, \text{1 month}, \text{1 year}\}$} as inputs, \added{whose durations are $d \in \{\text{1 hour}, \text{1 day}, \text{1 month}, \text{1 year}\}$, correspondingly.}
% % The data is processed across multiple resolutions 
% \deleted{where each resolution defines the granularity of the time series data and each duration determines the time span over which the input data is aggregated and processed.}

% Let $r$ represent the data resolution, where $r_0 = \text{5-min}$, $r_1 = \text{hourly}$, $r_2 = \text{daily}$, $r_3 = \text{monthly}$, and $r_4 = \text{yearly}$. 
The steady-state model can process time-series data with different resolutions $\{r_s\}$. The resolution progresses sequentially as follows:
\[
r_s \in \{\text{5-min} \rightarrow \text{hourly} \rightarrow \text{daily} \rightarrow \text{monthly} \rightarrow \text{yearly}\}.
\]
% For example, it can handle hourly data over a period of six years when $s=1$ and $e=4$. 
To accommodate steady-state data in different resolutions and durations, we employ a multi-layer structure.
% \replaced{multi-layer structure }{multiple feature extractors}. 
Each layer is designed to process data at a specific resolution \( r_s \) and transform it into a feature space aligned with the following higher resolution \( r_{s+1} \).
We denote the extractor handling data at resolution \( r_s \) as $\mathcal{M}_{r_s}(x_{r_s})$,
% \replaced{$\mathcal{M}_{r_s}(x_{r_s})$ }{\( \mathcal{M}_r \)},
where $x_{r_s}$ represents data from the $r_s$ resolution. Each encoding function \( \mathcal{M}_{r_s} \) takes an input $x_{r_s}$.
% \in\mathbb{R}^{n\times d}$ 
 The extractor then outputs a feature representation $z_{r_s}$ for the following resolution $r_{s+1}$ model processing.
% \replaced{for the next resolution $r_{s+1}$ model processing}{ for \( r_s \) data in \( r_{s+1} \) scale}.

\begin{figure*}[htbp]
    \centering
    % First plot with 0.7 linewidth
    % \begin{subfigure}[b]{0.5\linewidth}
    %     \centering
    %     \includegraphics[width=\linewidth]{plots/hierachical_model_arch.png}
    %     \caption{Hierarchical network for multiple time resolution and duration}
    %     \label{fig:network_overview}
    % \end{subfigure}
    % \hspace{0.05\linewidth}
    % \begin{subfigure}[b]{0.15\linewidth}
    %     \centering
    %     \includegraphics[width=\linewidth]{plots/pmu_model_arch.png}
    %     \caption{PMU network}
    %     \label{fig:pmu_model_arch}
    % \end{subfigure}
    % \hspace{0.05\linewidth}
    % % Second plot with 0.2 linewidth
    % \begin{subfigure}[b]{0.22\linewidth}
    %     \centering
    %     \includegraphics[width=\linewidth]{plots/feature_extractor_arch.png}
    %     \caption{Single feature extractor architecture}
    %     \label{fig:encoder_arch}
    % \end{subfigure}
        \includegraphics[width=0.8\linewidth]{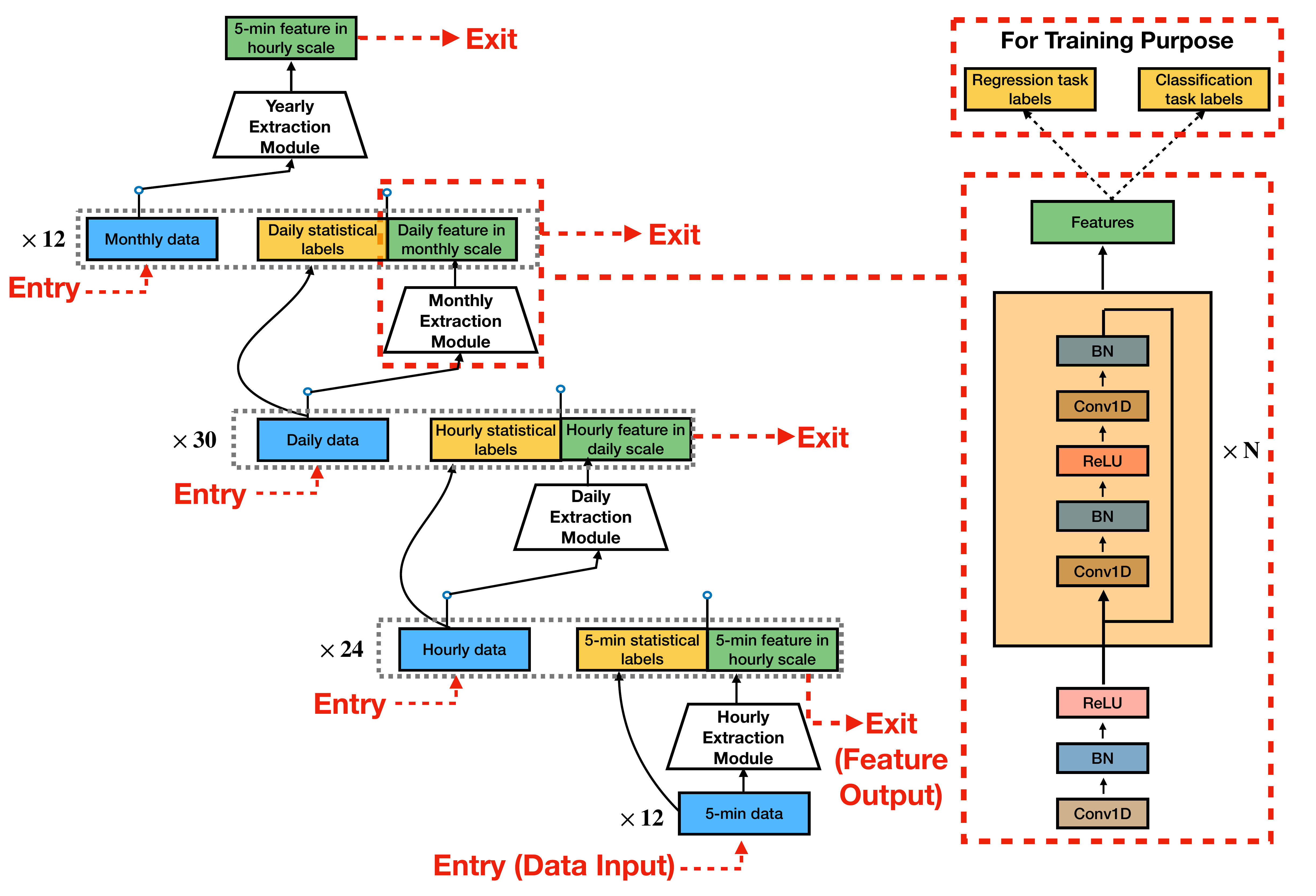}

    \caption{Overview of the hierarchical multi-resolution feature extraction framework, integrating multiple time-scale feature extractors to enhance synthetic data evaluation in smart grids.}
    \label{fig: Feature extraction model overview}
\end{figure*}

We employ these extractors at different resolutions in a hierarchical structure,
% where each level of the extractor processes data at a specific resolution and generates features that serve as input to the next level. 
where each subsequent extractor builds upon the features aggregated from the stack of extractors in the prior level. %Since only one resolution of raw data is needed during implementation, while raw data at all resolutions are available during training, 
A switch mechanism is introduced so that each extractor can adapt to the type of input it receives. Each extractor can select its input from either the features extracted at the prior resolution level, or the raw data at the current resolution level, 
% during implementation, or a combination of both during the training process, 
as demonstrated in Fig.~\ref{fig: Feature extraction model overview}. 
% Notably, each $x_{r_s}$ represents a unique dataset, which means $x_{r_s}$ and $x_{r_{s-1}}$ are treated as different datasets rather than the same dataset in different resolutions in our method 
% \textcolor{red}{I do not understand what this means}. 
For instance, the daily module (\( \mathcal{M}_{r_1} \)) can directly process 24 hourly data points if only hourly data is available, producing hourly features on a daily scale. Alternatively, if only 5-minute data is available, 24 hourly modules (\( \mathcal{M}_{r_0} \)) can be used to transform the 5-minute data for each hour (consisting of 12 data points per hour) into corresponding hourly features. Subsequently, the resulting 24 hourly features are concatenated and utilized as input to the daily module. 
% Additionally, the daily module can combine the stacked hourly features and the raw 24 hourly data points as inputs if both data are available. 
% \textcolor{red}{This needs clarification. "Both data are available"? What does it mean?}

% is optimally selected to minimize the difference between the feature and normal distributions. This optimization is crucial because the subsequent Fr\'{e}chet Distance calculation assumes that the features follow a normal distribution.
% \textcolor{red}{We need to use the word "optimization" very carefully. Did you actually perform some kind of search to optimize over it systematically, or did you just explore the choices and eyeball a good one?} 
% \replaced{This approach resolves a difficulty when combining raw data with feature representations: the feature representations are $D_{r_s}$ dimensional, whereas the raw time-series data are one dimensional. In this case, the higher-level data is viewed as the mean-value of the lower-level data values. The other lower-level extracted features can be simply zero-padded.}{To address this, we introduce an additional dimension to the input data at resolution $r_s$ and pad it to ensure alignment with the feature dimensions $D_{r_s}$ }{A challenge may arise when attempting to combine raw data with feature representations: the feature representations are $D_{r_s}$ dimensional, whereas the raw time-series data are one dimensional. To address this, we introduce an additional dimension to the input data at resolution $r_s$ and pad it to ensure alignment with the feature dimensions $D_{r_s}$.}

It is noteworthy that this structure has the capability to adapt to multi-resolution, multi-duration data. The input data can be injected at any designated \textit{entry} point in the model, depending on the data resolution \( r_s \). The model will process the data through a series of lower-level extractors to higher-level extractors and finally generate the desired feature representation at the module corresponding to the target scale $r_e$.

% However, a one-dimensional feature representation lacks sufficient descriptive power to capture the complexity of the data. To address this, we employ higher-dimensional feature vectors $ F_{r_s} $ Extracted from the module along with a statistical feature, precisely the mean value $\mu (X{r_{s-1}})$, which is directly computed from the data. We define $ D_{r_s} $ as the feature dimension associated with resolution $ r_s $. The dimension $D_{r_s} $ is optimally chosen based on the objective of making the feature distribution as close to a normal distribution as possible, which is essential for the subsequent Fr\'{e}chet Distance calculation. Since input data is one-dimensional time series, aligning it with the output of prior extractors is necessary. To achieve this, we pad the input data at resolution $r_s$ to match the feature dimensions $D_{r_s}$. 
% This padding ensures that the input size meets the requirements of each level of feature extractors within our hierarchical framework.

%  By implementing this setting, we can effectively 
% combine feature extractors in different resolutions and build a hierarchical structure that allows data in different resolutions to pass through multiple-level extractors to get its feature representation in the desired duration. 
% % It's a hierarchical model because a single input for the module in lower resolution is formed by output from multiple higher resolution modules so that more higher resolution modules are needed than lower resolution modules for a desired duration. 

During training and usage, the model processes the data in the form
$\left( \frac{N}{L_{r_s}}, D_{r_s}, L_{r_s} \right)$ given the resolution of data $r_s$.
Here, \( N \) denotes the total number of data points across all sequences.
Consider 1{,}000 data samples, each representing daily-scale data recorded at 5-minute intervals. Since one day has \( 288 \) such intervals, the total number of data points is $1{,}000 \times 288 = 288{,}000.$ \( L_{r_s} \) is the number of data points required at resolution \( r_s \) to form a single data point at the next higher resolution \( r_{s+1} \). For 5-min resolution data, $L_{\text{5-min}}=12$. Therefore, to process the entire batch of sequences, the dataset must be divided into segments of size \(\frac{N}{L_{r_s}}\). Under this setup, \(\frac{N}{L_{r_s}}\) modules $\mathcal{M}_{r_s}$ will each handle a segment of length \(L_{r_s}\), which aligns with the module design in our structure. Next, we define $D_{r_s}$ as the augmented feature dimension associated with resolution $r_s$. This dimension includes features extracted from the preceding module, as well as a one-dimensional mean value derived from the prior-level data. In contrast, the raw time-series data remain one-dimensional. In this case, the higher-level data are viewed as the mean value of the lower-level data values, and the other lower-level extracted features can be simply zero-padded. The dimension $D_{r_s}$ must be selected carefully to ensure that the FPD score computed in this feature space reliably and smoothly reflects the actual differences between datasets with different disturbance levels. To determine an appropriate dimension, we perform ablation tests by varying the feature dimension that achieves the best performance across different levels of Gaussian noise disturbance. An example ablation result for the hourly module is presented in SI Sec.S1; Fig. S1.
% in Fig.~\ref{fig:appA_feature_ablation} in the Appendix. 
The selected dimensions $D_{r_s}$ for all modules are summarized in SI Sec.S1; Table S4.
% Table~\ref {tab:appA_arch_scales} in the Appendix.

% approximately Gaussian distributed because the subsequent Fr\'{e}chet Distance calculation assumes that the features follow a Gaussian distribution. 

Mathematically, to extract feature vectors for input data $x_{r_s}$ at resolution $r_s$ over a target duration $r_e$, we define the feature extraction process as a sequential operation within the hierarchical framework. The feature output at each resolution can be expressed as: 
% Now assume we want to extract the feature vectors for input data $X_{r_s}$ with resolution $r_s$ on duration $r_e$ scale. We can achieve this by applying a sequentially approach within the hierarchical framework, which involves processing the data sequentially through feature extractors starting from the extractor $\mathcal{M}_{r_{s}}$, where each extractor maps features from one resolution to the next-level resolution until the desired duration $r_e$ is reached. In each iteration, the feature output can be calculated as:
\begin{equation}
    z_{r_s} = \mathcal{M}_{r_{s}}\left(\mathbf{I}_{r_s}\right),
\end{equation}
where $\mathbf{I}_{r_s}$ indicates the input at level $r_s$:
\[
\mathbf{I}_{r_s} =
\begin{cases}
\left[z_{r_{s-1}},\bar{x}_{r_{s-1}}\right],
& \text{if $z_{r_{s-1}}$ is available}\\ 
% &\\
\left[0,x_{r_s}\right], 
& \text{if $x_{r_{s}}$ is available} \\
% & \text{this level is used,} \\
% &\\
% \left[[\mathbf{z}_{r_{s-1}},\mu (x_{r_{s-1}})], [0,x_{r_s}]\right] 
% & \text{if $z_{r_{s-1}}$ and $x_{r_{s}}$ both}\\ 
% & \text{available (Training Process)}\\
\end{cases}
\]
In each resolution, we first determine input $\mathbf{I}_{r_s}$ between the feature representation from the prior level and the data at the current level. Due to the dimensional inconsistency between $\left[z_{r_{s-1}}, \bar{x}_{r_{s-1}}\right]$, which has dimension $D_{r_s}$, and $x_{r_s}$, which has dimension $1$, we need to pad $x_{r_s}$, if necessary, to match the dimension $D_{r_s}$. This operation ensures that all choices of $\mathbf{I}_{r_s}$ have the same dimension and can be adapted by the model. Also, considering the days in different months are inconsistent, for $r = 2$, we pad the length to 31 if needed. We then reshape $\mathbf{I}_{r_s}$ to size \( \left(\frac{N}{L_{r_s}}, D_{r_s}, L_{r_s}\right) \). This operation allows each module at a given resolution to process only a segment of the entire input sequence. Then, by stacking multiple such modules together, we ensure that the entire sequence in $r_e$ scale is effectively processed. The reshaped feature vector \( z_{r_s} \) is passed as input to the extractor at the resolution $r_{s+1}$. We then update the resolution to \( r_s = r_{s+1} \) and repeat this process sequentially until the target resolution $r_e$ is reached. Once $r_e$ is achieved, we output the resulting feature vectors, which can be used to calculate the FPD score.

% reshape \( X_{r_s} \) to the size \( \left(\frac{N}{L_{r_s}}, 1, L_{r_s}\right) \), This operation allows each module at a given resolution to process only a segment of the entire input sequence. Then, by stacking multiple such modules together, we ensure that the entire sequence with duration $d$ is effectively processed. Then we pad reshaped input to match the size \( \left(\frac{N}{L_{r_s}}, D_{r_s}, L_{r_s}\right) \). Next, \( X_{r_s} \) is provided as input to the corresponding module at resolution \( r_s \), which extracts a feature vector. The feature vector is concatenated with the mean of 
% \( X_{r_s} \) and reshaped to the appropriate size to match the input requirements of the next-resolution extractor. 

% During training, this combined input is used to form the input for the subsequent model. In implementation,

The parameters and modules corresponding to each resolution \( r \) are listed in SI Sec.S1; Table S4.
% Table \ref{tab:appA_arch_scales} in the Appendix. 
The feature extraction algorithm is outlined in Algorithm \ref{recursive_algorithm}.

\begin{algorithm}
\caption{Feature Extraction Algorithm}
\begin{algorithmic}[1]
\label{recursive_algorithm}

\STATE \textbf{Input:} Duration $r_e$, Start resolution $r_s$, Batch size $N$, Input data $x_{r_s}$, Feature dimension $D_{r_s}$, Data Length $L_{r_s}$ at resolution $r_s$  
\STATE \textbf{Output:} Final feature representation for data $x_{r_s}$ in $r_e$ scale
% $F_{r_e}$ at target resolution $r_e$
% \STATE Set $d = r_{\text{next}}$ if Training
% \STATE Set $r_e \gets \text{Mapping from duration } d$
% \STATE \textbf{Function FeatureExtraction}($X_{r_s}$, $r_s$, $r_e$)
\WHILE{$r_s < r_e$}
    \IF{$r_{s-1}$ is available}
        \STATE Set $\mathbf{I}_{r_s}$ to $\left[\mathbf{z}_{r_{s-1}},\bar{x}_{r_{s-1}}\right]$
    \ELSIF{$r_{s}$ is available}
        \STATE Set $\mathbf{I}_{r_s}$ to $[0,x_{r_s}]$
    % \ELSE
    % \STATE Set $\mathbf{I}_{r_s}$ to
    %     $\left[[\mathbf{z}_{r_{s-1}},\mu (x_{r_{s-1}})], [x_{r_i},0]\right]$
    \ENDIF
    \STATE Reshape ${I}_{r_s}$ to size $\left(\frac{N}{L_{r_s}}, D_{r_s}, L_{r_s}\right)$
    \STATE Calculate feature representation:     $z_{r_s} = \mathcal{M}_{r_{s}}\left(\mathbf{I}_{r_s}\right)$ 
    \STATE Advance resolution: $r_s = r_{s+1}$ 
\ENDWHILE
\STATE \textbf{Return} Output final feature vector $z_{r_s}$

% \STATE Reshape $X_{r_s}$ to dimensions $\left(\frac{N}{L_{r_s}}, 1, L_{r_s}\right)$
% \STATE Pad $X_{r_s}$ with zeros to match the size $\left(\frac{N}{L_{r_s}}, D_{r_s}+1, L_{r_s}\right)$
%     \IF{$r_s == r_e$} 
%         \STATE Apply module at resolution $r_s$: $F_{r_s} \gets \mathcal{M}_{r_s}(X_{r_s})$
%         \RETURN $F_{r_s}$
%     \ELSE
%         \STATE Get next resolution: $r_{\text{next}}$ based on $r_s$
%         \STATE Compute feature output at current resolution: $F_{r_s} \gets \mathcal{M}_{r_s}(X_{r_s})$
%         \STATE Compute mean of input data: $\mu(X_{r_s}) \gets \frac{1}{L_{r_s}} \sum_{t=1}^{L_{r_s}} X_{r_s}[:, t]$
%         \STATE Concatenate features with mean: $F_{r_s} \gets [F_{r_s}, \mu(X_{r_s})]$
%         \STATE Reshape $F_{r_s}$ to size $(\frac{N}{L_{r_\text{next}}},D_{r_\text{next}}+1,L_{r_\text{next}})$
%         \IF {Training}
%             \STATE Padding $X_{r_{\text{next}}}$ with 0
%             \STATE Concatenate updated features with padded $X_{r_{\text{next}}}$: 
%             $X_{r_{\text{next}}}\gets [[F_{r_s}], [X_{r_{\text{next}}}]]$ 
%         \ELSE
%             \STATE Update $X_{r_{\text{next}}}$ with features:
%             $X_{r_{\text{next}}}\gets F_{r_s}$
%         \ENDIF
%         \STATE \textbf{return} FeatureExtraction($X_{r_{\text{next}}}$, $r_{\text{next}}$, $r_e$)
%     \ENDIF
% \STATE \textbf{End Function}

% \STATE $F_{r_e} \gets \text{FeatureExtraction}(X_{r_s}, r_s, r_e)$

% \STATE \textbf{Return} $F_{r_e}$

\end{algorithmic}
\end{algorithm}
The architecture designs are simpler in the transient-state cases. For this case, we use a single-time resolution with a 120Hz sampling frequency. Hence, we only apply a single feature extractor to the transient-state related task. A data sample is an 8-second length sequence with three channels of voltage magnitude, phase angle, and frequency, all represented in per-unit form. The input is first upsampled to 32 channels and finally downsampled by stride of 2 filters to a feature vector with 2048 dimensions. Then, we can calculate the FPD scores based on the extracted feature vectors.

For the architecture design of a single feature extractor in both models, we utilize residual blocks in the feature extractor as demonstrated in Fig.~\ref{fig: Feature extraction model overview}. The residual connections have been widely studied and proven to increase neural network depth to enhance representation capabilities. 
% The input dimensions of time-series data are $(\frac{N}{L_r},D_r,L_r)$ after expanding to align with the dimension of the feature space. 
We utilize the \textit{conv-1d} and \textit{batchnorm-1d} layers for accommodating the time-series inputs. This ResNet-like feature extractor is a base component for different scenarios and timescale tasks, followed by simple linear heads for different downstream training objectives.

% \begin{figure*}[htbp]
%     \centering
%     % First plot with 0.7 linewidth
%     \begin{subfigure}[b]{0.7\linewidth}
%         \centering
%         \includegraphics[width=\linewidth]{plots/Network Setting Overview.pdf}
%         \caption{Hierarchy Network Overview}
%         \label{fig:network_overview}
%     \end{subfigure}
%     \hspace{0.01\linewidth}
%     % Second plot with 0.2 linewidth
%     \begin{subfigure}[b]{0.2\linewidth}
%         \centering
%         \vspace{-1cm}
%         \includegraphics[width=\linewidth]{plots/encoder_architecture.png}
%         \vspace{1cm}
%         \caption{Single feature extractor architecture}
%         \label{fig:encoder_arch}
%     \end{subfigure}

%     \caption{Feature extraction model overview}
%     \label{fig:two_plots}
% \end{figure*}

\subsection{Fr\'{e}chet Power-Scenario Distance (FPD) calculation}
Once robust representations are obtained from the feature extractor, we can then compare the distribution similarity between the two datasets in the feature space. 
% \textcolor{red}{Tian: Where was this demonstrated? In previous sections, we demonstrated why we want to use distribution evaluation metrics on extracted feature space over other method metrics.} 
Three distribution-level evaluation metrics are commonly used\cite{razghandi2022variational}: (1) Jensen–Shannon (JS) Divergence, (2) Kernel Distance (Maximum Mean Discrepancy - MMD), and (3) Fr\'{e}chet Distance (Wasserstein-2 Distance).

% The features can often be approximately viewed as having a Gaussian distribution. 
Denote the feature vectors extracted from two sets $x_r^1$ and $x_r^2$ at resolution $ r $ as $z_r^1$ and $z_r^2$. The means of extracted features can then be expressed as:
% of inputs $x_r$ and $y_r$ at resolution $ r $ as:
\begin{eqnarray}
\mathbf{m}_1 = \frac{1}{N} \sum_{i=1}^{N} z_{r,i}^1, \quad \mathbf{m}_2 = \frac{1}{N} \sum_{i=1}^{N} z_{r,i}^2
\end{eqnarray}
The covariance matrices of the feature vectors extracted are:
\begin{eqnarray}
\Sigma_1 = \frac{1}{N} \sum_{i=1}^{N} \left( z_{r,i}^1 - \mathbf{m}_1 \right)\left( z_{r,i}^1 - \mathbf{m}_1 \right)^T
\\
%\]
%\[
\Sigma_2 = \frac{1}{N} \sum_{i=1}^{N} \left( z_{r,i}^2 - \mathbf{m}_2 \right)\left( z_{r,i}^2 - \mathbf{m}_2 \right)^T
\end{eqnarray}

Using the Jensen–Shannon (JS) divergence—the symmetrized and smoothed variant of KL \cite{nielsen2019jensen} —the similarity between $z_r^1$ and $z_r^2$ is defined as
% \begin{equation}
% \mathrm{JS}(z_r^1 \,\|\, z_r^2) \;=\; \tfrac{1}{2}\,\mathrm{KL}\!\left(\mathcal{N}(\mathbf{m}_1,\Sigma_1)\,\Big\|\,\mathcal{M}\right)
% \;+\; \tfrac{1}{2}\,\mathrm{KL}\!\left(\mathcal{N}(\mathbf{m}_2,\Sigma_2)\,\Big\|\,\mathcal{M}\right),
% \end{equation}
\begin{equation}
\begin{aligned}
\mathrm{JS}(z_r^1 \,\|\, z_r^2)
&= \tfrac{1}{2}\,\mathrm{KL}\!\big(\mathcal{N}(\mathbf{m}_1,\Sigma_1)\,\big\|\,\mathcal{N}(\mathcal{M}\big) \\
&\quad + \tfrac{1}{2}\,\mathrm{KL}\!\big(\mathcal{N}(\mathbf{m}_2,\Sigma_2)\,\big\|\,\mathcal{N}(\mathcal{M}\big).
\end{aligned}
\end{equation}
% \[
% \mathbf{m}_M=\tfrac{1}{2}(\mathbf{m}_1+\mathbf{m}_2), \qquad \Sigma_M=\tfrac{1}{2}(\Sigma_1+\Sigma_2).
% \]
where $\mathcal{M}$ is the (Gaussian) mixture approximation with mean and covariance
$\mathbf{m}_M=\tfrac{1}{2}(\mathbf{m}_1+\mathbf{m}_2)$ and $\Sigma_M=\tfrac{1}{2}(\Sigma_1+\Sigma_2)$, and
$\mathrm{KL}(\cdot\|\cdot)$ denotes the closed-form KL divergence between Gaussians:
% \begin{equation}
% \mathrm{KL}\!\left(\mathcal{N}(\mathbf{m}_a,\Sigma_a)\,\Big\|\,\mathcal{N}(\mathbf{m}_b,\Sigma_b)\right)
% =\tfrac{1}{2}\!\left[\mathrm{Tr}(\Sigma_b^{-1}\Sigma_a)+(\mathbf{m}_b-\mathbf{m}_a)^\top \Sigma_b^{-1}(\mathbf{m}_b-\mathbf{m}_a)-d+\log\!\frac{\det\Sigma_b}{\det\Sigma_a}\right].
% \end{equation}
\begin{equation}
\begin{aligned}
&\mathrm{KL}\!\left(\mathcal{N}(\mathbf{m}_1,\Sigma_1)\,\middle\|\,\mathcal{N}(\mathbf{m}_2,\Sigma_2)\right)
= \tfrac{1}{2}\Big[\operatorname{tr}(\Sigma_2^{-1}\Sigma_1)
+ \\
&(\mathbf{m}_2-\mathbf{m}_1)^\top \Sigma_2^{-1}(\mathbf{m}_2-\mathbf{m}_1) 
- d + \log\!\frac{\det \Sigma_2}{\det \Sigma_1}\Big].
\end{aligned}
\end{equation}
Here $d$ is the dimensionality of the feature space for $z_r^1$ and $z_r^2$. 
JS divergence is symmetric and bounded, offering improved stability over plain KL. However, when the distributions have little or no support overlap, JS becomes insensitive—it can saturate at its maximum and fail to track changes in similarity reliably, as illustrated in \cite{arjovsky2017wasserstein}. 
% like KL it still requires the two representations to have the \emph{same} dimensionality. 
% Because our goal is to compare datasets across resolutions (i.e., different feature dimensions), 
% Therefore, JS divergence is likewise not suitable for our purpose.

 % \textcolor{red}{?? d is a single number, how to represent two values?}

With the Kernel distance as the evaluation metric, the distance between $z_r^1$ and $z_r^2$ is calculated as:
\begin{equation}
% \begin{aligned}
% \text{MMD}(X_r, Y_r) = \\\|\mathbf{m}_1 - \mathbf{m}_2\|^2 + \frac{1}{N^2} \sum_{i=1}^{N} \sum_{j=1}^{N} \left( k(F_r(X_{r,i}), F_r(X_{r,j})) \\+ k(F_r(Y_{r,i}), F_r(Y_{r,j})) - 2 k(F_r(X_{r,i}), F_r(Y_{r,j})) \right)
% \end{aligned}
\begin{aligned}
&\text{MMD}(z_r^1, z_r^2) =
\\&\; \|\mathbf{m}_1 - \mathbf{m}_2\|^2 + \frac{1}{N^2} \sum_{i=1}^{N} \sum_{j=1}^{N} \Big(k(z_{r,i}^1, z_{r,j}^1)\\&
 \; + k(z_{r,i}^2, z_{r,j}^2) - 2 k(z_{r,i}^1, z_{r,j}^2) \Big),
\end{aligned}
\end{equation}
where \( k(x, y) \) is a kernel function (e.g., Gaussian kernel). Choosing a proper kernel is extremely important for the metric to be effective, as different kernel functions can lead to varying metric performances.

We use the Fr\'{e}chet distance due to its empirically observed stability and consistency. Since this metric is specifically tailored to evaluate power system scenarios, we refer to this as Fr\'{e}chet Power-Scenario Distance (FPD).
The $ \text{FPD}(z_{r}^1,z_{r}^2) $ between the extracted features at resolution $ r $ is calculated as: 
% \textcolor{red}{NEVER start a new sentence with "And"!!!}
\begin{equation}
\label{equation:FPD}
 \text{FPD}(z_{r}^1,z_{r}^2) = \|\mathbf{m}_1 - \mathbf{m}_2\|^2 + \text{Tr}(\Sigma_1 + \Sigma_2 - 2(\Sigma_1 \Sigma_2)^{1/2})   
\end{equation}
FPD is essentially a form of the Fr\'{e}chet distance, where we assume that the features of the two datasets follow Gaussian distributions\footnote{For practical reasons, we consider only the mean and variance of distributions. Since the Gaussian maximizes entropy subject to fixed mean and covariance, we model feature distributions as multivariate Gaussian.}. This metric quantifies the similarity between the two distributions: a smaller FPD value indicates a more significant similarity between the feature distributions, ultimately reflecting a higher similarity between the datasets.

\section{Model Training}
\label{section:training}
% \begin{table}[ht]
% \centering
% \caption{Regression labels for steady-state training}
% \label{tab:regression_label}
% \begin{tabular}{l}
% \hline
% \textbf{Regression Tasks} \\
% \hline
% Minimum \\
% Maximum \\
% Range \\
% Variance \\
% Number of data points above average \\
% Number of data points below average \\
% Skewness \\
% Kurtosis \\
% Coefficient of variation \\
% \hline
% \end{tabular}
% \end{table}

% \begin{table}[h]
% \centering
% \caption{Hyper-parameters for steady-state model training }
% \centering
% \scalebox{1.0}{
% \centering
% \begin{tabular}{lc}
% \toprule
% Parameter & Setting \\
% \midrule
% Minibatch size & 1024 \\
% Optimizer & Adam \\
% Optimizer: learning rate & 1e-3 \\

% % Optimizer: weight decay & 2e-5 \\
% % Classification loss coefficient & 1.0 \\
% % Regression loss coefficient & 1.0 \\
% \bottomrule
% \end{tabular}
% }
% \label{tab:steady_param}
% \end{table}

% \begin{table}[h]
% \centering
% \caption{Hyper-parameters for transient-state model training }
% \centering
% \scalebox{1.0}{
% \centering
% \begin{tabular}{lc}
% \toprule
% Parameter & Setting \\
% \midrule
% Minibatch size & 128 \\
% Optimizer & Adam \\
% Optimizer: learning rate & 1e-4 \\
% Optimizer: weight decay & 2e-5 \\
% Classification loss coefficient & 1.0 \\
% Regression loss coefficient & 1.0 \\
% \bottomrule
% \end{tabular}
% }
% \label{tab:param}
% \end{table}
The purpose of training is to guide the feature extraction model to produce features that are tightly coupled to the smart grid datasets while retaining strong representational power.
The training data used in this study consists of multiple open-source datasets, each capturing different generative tasks. These datasets include various types of demand, EV charging profiles, and various types of electricity generation. Each dataset varies in terms of resolution and timescale, contributing to a comprehensive representation of the smart grid domain. The specific characteristics of each dataset are summarized in SI Sec.S3; Table S8.

For the steady-state model, each module in the feature extraction model is trained individually and sequentially from the bottom level (5-minute module) to the top level (yearly module). Once a certain level module has finished training, the parameters will be fixed, and we will start to train the next level module. The training data for the module at each resolution is $I_{r_s}$.
% = \left[[\mathbf{z}_{r_{s-1}},\mu (x_{r_{s-1}})], [x_{r_i},0]\right]$ 
% where 5-min module only input $I_{r_0} = x_{r_0}$. 
The training dataset is shown in SI Sec.S3; Table S8.
% Table \ref{tab:appC_datasets} in the Appendix.

Each module is trained based on the loss function:
\begin{equation}
\label{loss_function}
    \begin{aligned}
    L_{r_s} = \frac{1}{N} \sum_{i=1}^N \Bigg[ & \sum_{k=1}^K\left( f_{\text{reg}}^k(I_{r_s}) - y_i^{\text{reg}} \right)^2 \\
    & - \sum\limits_{i=1}^C y_i^{\text{clf}}\log \left( 
    \text{softmax}(f_{\text{clf}}(I_{r_s}))\right) \Bigg]
    \end{aligned}
\end{equation}

% Both $f_{\text{reg}}$ and $f_{\text{clf}}$ are Multi-Layer Perceptrons (MLPs) that take the feature outputs from the module and generate the corresponding task predictions for training purposes. Here, each MLP consists of multiple layers of neurons that are fully connected:

Both $f_{\text{reg}}$ and $f_{\text{clf}}$ are fully connected layers that take the feature output from the module and generate the corresponding predictions of tasks. Here, $W$ denotes the layer weights, and $b$ denotes the layer biases:

\begin{equation}
f_{\text{reg}}^k(I_{r_s}) = W_{\text{reg}}^k \mathcal{M}(I_{r_s}) + b_{\text{reg}}^k
\end{equation}

\begin{equation}
    f_{\text{clf}}(I_{r_s}) = W_{\text{clf}} \mathcal{M}(I_{r_s}) + b_{\text{clf}}
\end{equation}

The first term in (\ref{loss_function}), $\sum_{k=1}^K\left( f_{\text{reg}}^k(I_{r_s}) - y_i^{\text{reg}} \right)^2$, corresponds to the Mean Squared Error (MSE) loss across a total of $K=9$ regression tasks. 
Regression tasks are designed for
leading extractors to capture features that
reflect general statistical characteristics of data represented by $y_i^{\text{reg}}$ that transfer across signal types (e.g., rising or falling patterns), enabling the model to extract similar global representations when signals share statistical behavior.
% reflect a specific statistical characteristic of the data, represented by $y_i^{\text{reg}}$. 
The list of regression tasks is shown in SI Sec.S1; Table S1.
% Table \ref{tab:appA_regression_labels} in the Appendix. 
The second term,$- \sum\limits_{i=1}^Cy_i^{\text{clf}}\log \left( 
    \text{softmax}(f_{\text{clf}}(I_{r_s}))\right)$, represents the Cross-Entropy Loss for the classification task, which categorizes data based on classification labels $y_i^{\text{clf}}$.
    Classification tasks are designed to enable the extractors to capture local dynamics tied to the data source for smart grids (e.g., solar power is always zero at night), capturing signal-specific features.
    $y_i^{\text{clf}}$ is determined by the dataset index shown in SI Sec.S3; Table S7
    % Table \ref{tab:appC_datasets} 
    and $C$ is the number of categories of the dataset. The hyper-parameters for training are shown in SI Sec.S1; Table S2.
    % Table \ref{tab:appA_steady_hparams} in the Appendix.

% As outlined in the previous section and based on Algorithm \ref{recursive_algorithm}, despite the transient-state model, other models are trained sequentially from high to low resolution. Let the current model be $\text{Model}_{r_{s}}$ with resolution $r_{s}$, and the previous resolution be $r_{prev}$. During training, the data corresponding to the current resolution, $X_{r_{s}}$, along with the features extracted from the previous resolution, $F_{r_{prev}}$, and the mean of the previous layer's data, $\mu (X_{r_{prev}})$, are used. After training each model, its parameters are fixed, and we start to proceed with the model of the following resolution. 

% Training for each model integrates a classification and a regression task \textcolor{red}{Only two tasks here? We also need a formal definition of the training cost function. Some details on the training setup, optimizer, hardware-software environment, training time, etc. need to be included}. The classification task is designed to categorize data based on dependencies, while the regression task guides the feature extractor to capture features that reflect the data's statistical characteristics.

% categorizes the data into one of nine predefined categories. In contrast, the regression task trains the model to match nine selected statistical characteristics of the data.
% shown in appendix \ref{app:hier}.

For transient-state model training, we use fault type, minimum amplitude, and maximum amplitude as supervised pretraining labels. We then exploit the pretrained feature extractor to evaluate synthetic samples. The length of a data sequence is 960, and the input data is encoded into a 2048-dimensional representation. Moreover, the hyper-parameters for training are shown in SI Sec.S1; Table S3.
% Table \ref{tab:appA_transient_hparams} in the Appendix.
Once training is complete, we discard the task heads and use the trained module $\mathcal{M}(I_{r_s})$ solely for feature extraction. 

All training and experiments were conducted on a 16 GB Apple M2 chip MacBook Pro. The model was trained using PyTorch 2.2, leveraging Apple's Metal Performance Shaders (MPS) backend for GPU acceleration.

%These outputs serve as the features extracted, which can then be used to compute the FPD using the equation \ref{equation:FPD}

% \begin{table}[ht]
% \centering
% \caption{Regression labels for steady-state training}
% \label{tab:regression_label}
% \begin{tabular}{l}
% \hline
% \textbf{Regression Tasks} \\
% \hline
% Minimum \\
% Maximum \\
% Range \\
% Variance \\
% Number of data points above average \\
% Number of data points below average \\
% Skewness \\
% Kurtosis \\
% Coefficient of variation \\
% \hline
% \end{tabular}
% \end{table}

% \begin{table}[h]
% \centering
% \caption{Hyper-parameters for steady-state model training }
% \centering
% \scalebox{1.0}{
% \centering
% \begin{tabular}{lc}
% \toprule
% Parameter & Setting \\
% \midrule
% Minibatch size & 1024 \\
% Optimizer & Adam \\
% Optimizer: learning rate & 1e-3 \\

% % Optimizer: weight decay & 2e-5 \\
% % Classification loss coefficient & 1.0 \\
% % Regression loss coefficient & 1.0 \\
% \bottomrule
% \end{tabular}
% }
% \label{tab:steady_param}
% \end{table}

% \begin{table}[h]
% \centering
% \caption{Hyper-parameters for transient-state model training }
% \centering
% \scalebox{1.0}{
% \centering
% \begin{tabular}{lc}
% \toprule
% Parameter & Setting \\
% \midrule
% Minibatch size & 128 \\
% Optimizer & Adam \\
% Optimizer: learning rate & 1e-4 \\
% Optimizer: weight decay & 2e-5 \\
% Classification loss coefficient & 1.0 \\
% Regression loss coefficient & 1.0 \\
% \bottomrule
% \end{tabular}
% }
% \label{tab:param}
% \end{table}

\section{Effectiveness Evaluation of the FPD} 
\label{section:validation}

\subsection{Power-Aware Feature Extraction vs. Generic Encoders}
In this section, we demonstrate that the proposed extractor produces representations that align more closely with these physical principles than generic time-series encoders (e.g., TS2Vec \cite{yue2022ts2vec}, TimeMixer \cite{wang2024timemixer}) on solar generation data. For each extractor/encoder, we compute similarity using the same Fr\'{e}chet distance in the respective feature space:
\begin{itemize}
  \item \textbf{Period Offset}: Power signals exhibit strong periodicity (e.g., daily/seasonal load patterns; solar generation following the diurnal cycle—zero at night, peak near noon). To test sensitivity to disturbed periodic structure, we apply a time offset of the solar trace by \(\alpha \in \{0,2,4\}\) hours. Here, \(\alpha\) quantifies how far the shifted solar profile is displaced from its true period. A robust extractor/encoder should distinguish these shifted variants from the nominal series.

  \item \textbf{Nighttime Generation (Physical Violation)}: Power signals must respect physical constraints (e.g., bounded ramp rates; no solar output at night). We inject nonzero solar generation during nighttime to create a controlled violation and evaluate whether extractors/encoders can detect it. The violation level is parameterized by \(\alpha \in \{0,2,3\}\), where \(\alpha\) denotes the number of nighttime hours forced to have nonzero solar output.

  \item \textbf{Cross–Time-Resolution Consistency}: A key property of power signals is that the same underlying process may be represented at different sampling resolutions; representations derived from hourly and 5-minute data should remain consistent. Our feature extractor employs a hierarchical, multi-scale design to capture structure across resolutions, whereas generic encoders (e.g., single-scale inputs) may treat different resolutions as unrelated sequences. To demonstrate this advantage, we compare features from hourly solar/wind datasets against their 5-minute counterparts and assess their similarity.
\end{itemize}

\begin{figure}[t]
    \centering
    \includegraphics[width=1\linewidth]{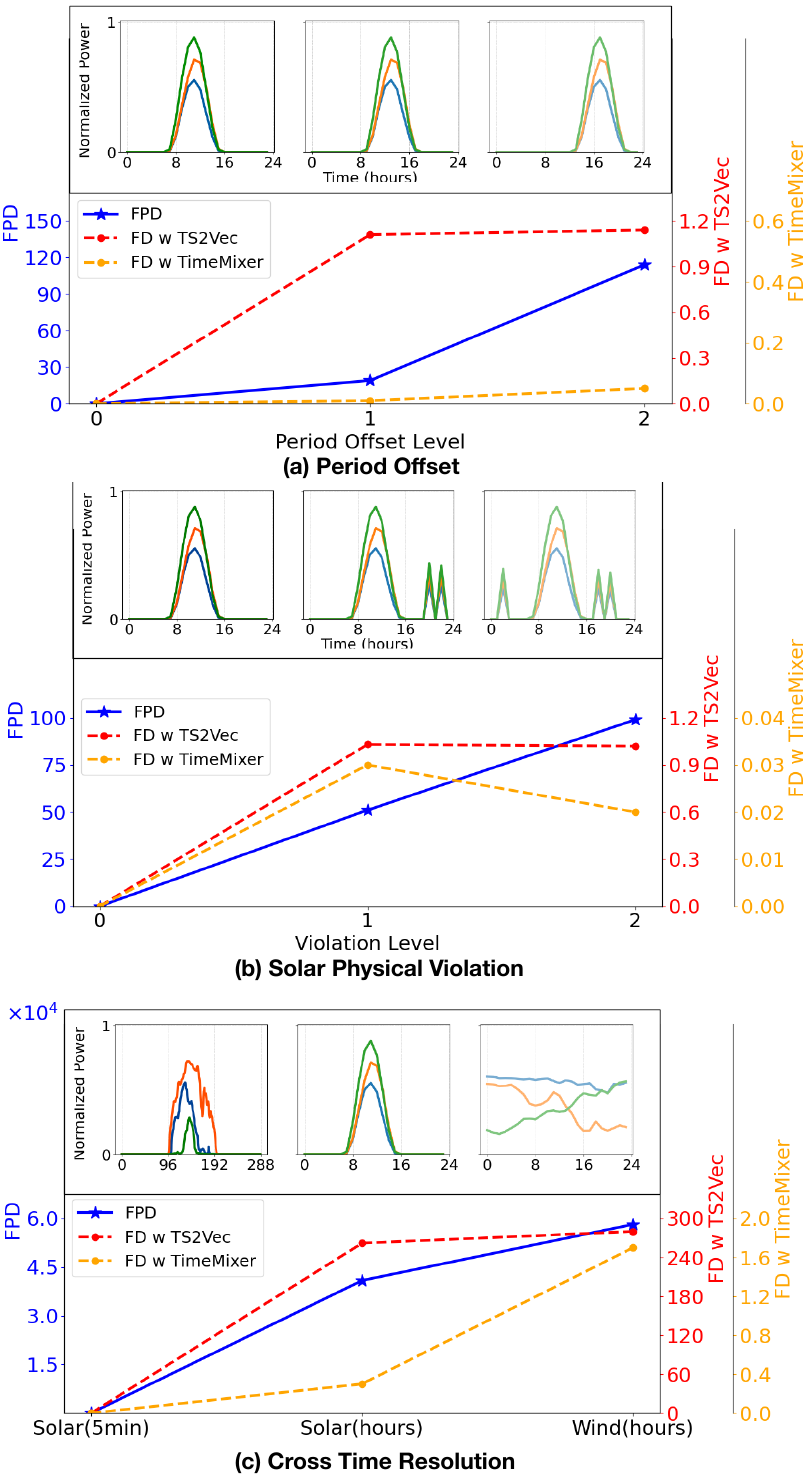}
    \caption{Features from the proposed power-aware extractor vs. generic time-series encoders, evaluated using the Fr\'{e}chet distance in feature space in case: 
\textbf{(a) Period Offset}: solar series are shifted by 
\(\alpha \in \{0,2,4\}\) hours. 
\textbf{(b) Solar Physical Violation}: Unrealistic power generation is injected for
\(\alpha \in \{0,2,3\}\) hours of solar generation during night. 
\textbf{(c) Cross Time Resolution}: 5-minute solar compared against hourly solar/wind.}
    \label{fig:FPD_extractors}
\end{figure}

As shown in Fig.~\ref{fig:FPD_extractors}, the proposed extractor consistently and effectively distinguishes period shifts, detects physical feasibility violations, and preserves similarity judgments across time resolutions. By contrast, features from generic encoders do not reliably reveal phase misalignment or feasibility violations and yield inconsistent cross-resolution similarity, indicating reduced sensitivity to power-specific structure.

These findings demonstrate that our extractor is better aligned with power-system characteristics than generic models. By faithfully capturing the distinct temporal regularities, physical constraints, and multi-resolution structure of power data, it provides a solid foundation for more reliable evaluation and analysis in subsequent sections.

\subsection{Distributional Metric Suitability in Learned Feature Space}
\label{section:effectiveness}

\begin{figure*}[htbp]
    \centering
        \includegraphics[width=1.0\linewidth]{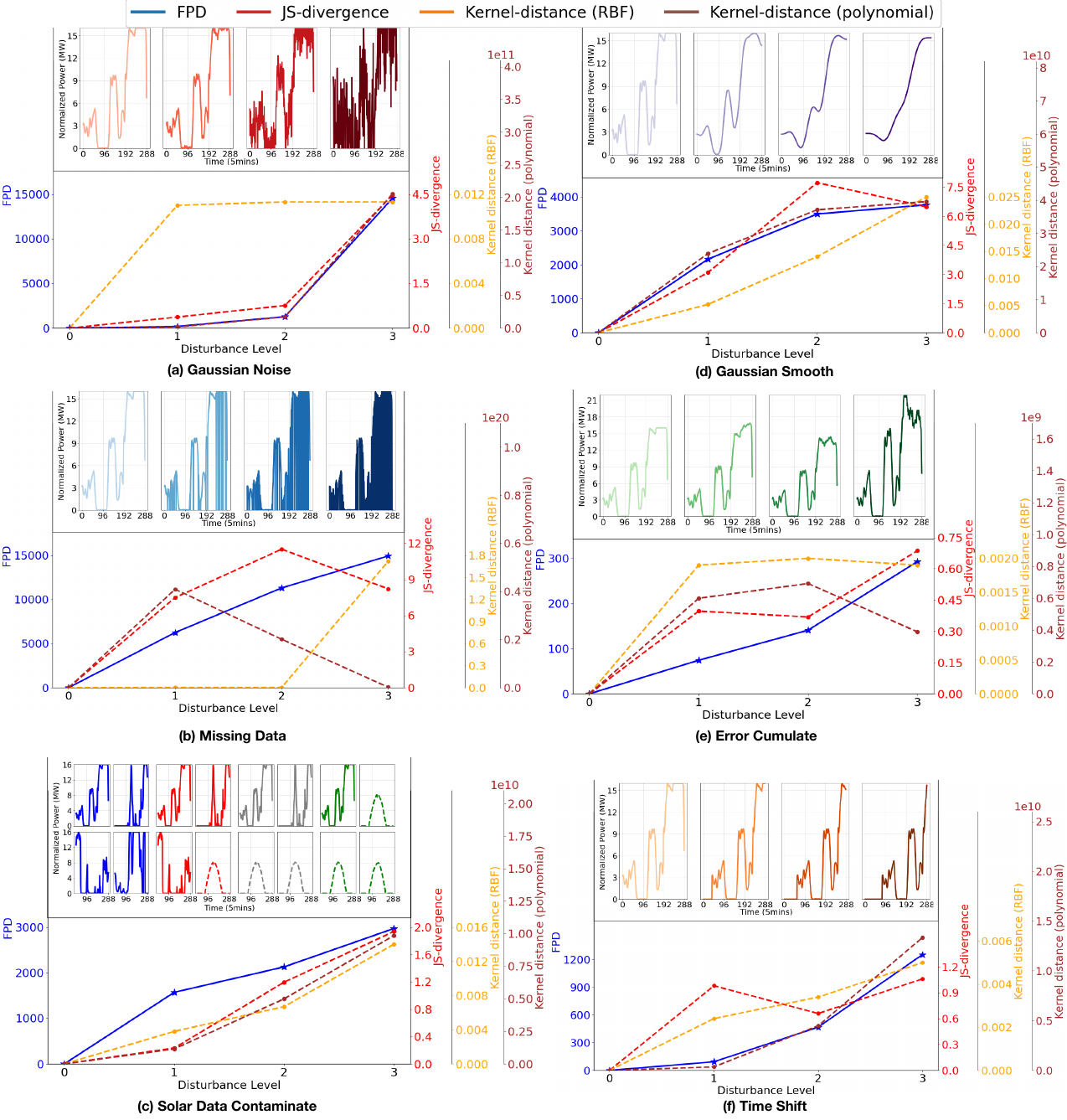}
\caption{
Effectiveness of FPD with six different types of disturbances on the dataset \(X\):
% (\ref{fig:Gaussian Noise}) 
\textbf{(a) Gaussian Noise} is added with mean 0 and varying variances \(\alpha = [0, 0.16, 1.6, 4]\), where a larger \(\alpha\) indicates more noise.
% (\ref{fig:Missing Data}) 
\textbf{(b) Missing Data} is simulated by randomly replacing \(\alpha \times 100\%\) of the points in each sample of \(X\) with zeros, where \(\alpha = [0, 0.1, 0.25, 0.5]\).
% (\ref{fig:Salt-and-Pepper Noise}) \textbf{Salt-and-Pepper Noise} is introduced by randomly selecting \(\alpha \times 100\%\) of the points in \(X\) and setting each to either the minimum or maximum value, with \(\alpha = [0, 0.02, 0.04, 0.06]\).
% (\ref{fig:Contaminate}) 
\textbf{(c) Solar Data Contamination} replaces \(\alpha \times 100\%\) of the points in \(X\) with corresponding points from a solar power dataset \(Y\)\cite{nrel_sind_toolkit}, where \(\alpha = [0, 0.25, 0.5, 0.75]\).
% (\ref{fig:Blur}) 
\textbf{(d) Gaussian Smooth} applies Gaussian filter which has variance $ \alpha = [0, 10, 20, 30]$ to the original dataset. A larger $\alpha$ indicates more significant smoothing.
% (\ref{fig:Cumulate}) 
\textbf{(e) Error Cumulate} is simulated by adding an error, which is multipled by a random Gaussian noise with a mean of 1 and varying variances \(\alpha = [0, 0.005, 0.01, 0.03]\) at each time step.
% (\ref{fig:shift}) 
\textbf{(f) Time Shift} is a potential problem that exists in time-series generation. To simulate this effect, we shift the data forward by $\alpha = [0,40,60,80]$ intervals.
}
    \label{fig:FPD Performance}
\end{figure*}
In the previous section, we showed that the learned feature space produced by our extractor aligns with key smart grid properties. Here, we further present experiments that motivate the choice of the Fr\'{e}chet distance over alternative distributional metrics for measuring similarity within this learned space. The experiments are designed based on the Wind Generation Dataset (Dataset 1 in SI Sec.S3; Table S8)
% Table \ref{tab:appC_datasets}). 
We progressively increase the disturbance applied to the original dataset X in each experiment to provide a controllable and scalable measure of FPD performance.

\begin{itemize}
    \item \textbf{Gaussian Noise}: In this case, Gaussian noise with a mean of 0 and a variance \(\alpha\) is added to \(X\), where \(\alpha \in [0, 0.16, 1.6, 4]\) denotes the disturbance level. A larger \(\alpha\) indicates more noise added.  
    
    \item \textbf{Missing Data}: To simulate missing data points in \(X\), we randomly replace some values with zeros for each sample in this case. The disturbance \(\alpha \in [0, 0.1, 0.25, 0.5]\) indicates that \(\alpha \times 100\%\) of the points in \(X\) are randomly set to zero.
    
    % \item \textbf{Salt-and-Pepper Noise}: In this case, salt-pepper noise is added to \(X\). The disturbance $\alpha \in [0, 0.02, 0.04, 0.06]$ means that \(\alpha \times 100\%\) of the points in each sample of \(X\) are randomly selected. Each selected point has a 50\% chance of being set to either the minimum or maximum value of \(X\).
    
    \item \textbf{Solar Data Contamination}: To test cross-dataset discrimination, we contaminate the wind dataset \(X\) with samples from the solar dataset \(Y\) \cite{nrel_sind_toolkit}. The disturbance \(\alpha \in [0, 0.25, 0.5, 0.75]\) indicates that \(\alpha \times 100\%\) of the samples in \(X\) are replaced by corresponding points from \(Y\).

    \item \textbf{Gaussian Smooth}: To simulate scenarios where the generative model may fail to capture fine details of the original data, resulting in overly smooth outputs, we intentionally apply a Gaussian filter to blur these details. This operation allows us to test whether the FPD metric can detect such changes. The disturbance parameter \(\alpha \in [0, 10, 20, 30]\) represents the standard deviation of the Gaussian filter applied to the original data, with higher values of $\alpha$ indicating more robust smoothing.

\item \textbf{Error Accumulation}:  
Certain generative models, like Recurrent Neural Networks (RNNs), struggle with long sequences due to cumulative error over time. We introduce an accumulating error by applying Gaussian noise at each time step to simulate this effect. Specifically, for each time step \( t \) in sequence length \( T \), we add a noise factor \(\epsilon_t^{(\alpha)} \sim \mathcal{N}(1, \alpha)\) which is sampled from a Gaussian distribution centered at 1 with varying standard deviations, \(\alpha = [0, 0.005, 0.01, 0.03]\). For each time step, we compute cumulative modification factors \(E^{(\alpha)}_t\) recursively:
\[
E^{(\alpha)}_t = E^{(\alpha)}_{t-1} \cdot \epsilon_t^{(\alpha)}, \quad E^{(\alpha)}_0 = 1
\]
The modified error cumulated data \(\tilde{x}^{(\alpha)}\) is then given by:
\[
\tilde{x}^{(\alpha)}_t = x_t \cdot E^{(\alpha)}_t
\]
where \(x_t\) is the original data at time \(t\). We demonstrate that the FPD metric effectively detects and highlights this error accumulation effect.
%     \item \textbf{Error Cumulate}: 
% Some generative models, such as Recurrent Neural Networks (RNNs), perform poorly on long sequences due to error accumulation over time. To simulate this effect, we introduce an accumulating error by adding Gaussian noise at each time step, with a mean of 1 and varying standard deviation \(\alpha = [0, 0.005, 0.01, 0.03]\) at each time step. We then demonstrate that the FPD metric effectively highlights this accumulated error.

    \item\textbf{Time Shift}: Time shift in time-series data is a common issue in generative models, often leading to misalignment of temporal features between the original and generated data. This misalignment can be particularly problematic in smart grid data, where certain temporal features, such as peaks, carry critical information. To simulate this, we shift the data forward by intervals of $\alpha = [0, 40, 60, 80]$, creating misleading temporal dependencies. Our results demonstrate that FPD effectively detects and captures these temporal shifts.
\end{itemize}

% As shown in Fig. \ref{fig:FPD Performance}, KL-divergence can sometimes fail due to its asymmetry feature. The direction of comparison may lead to different penalty preferences, resulting in inconsistent outcomes. Kernel Distance performs well in some cases but may also fail to capture significant changes as disturbance levels vary. Additionally, the results of Kernel Distance are susceptible to the choice of kernel type and its parameters, which can lead to inconsistent results. It is also important to note that the scale of both methods can vary significantly across different disturbance scenarios.
% In contrast, Fr\'{e}chet Distance consistently handles all disturbances and effectively reflects changes in disturbance levels, indicating its reliability across all scenarios. Moreover, Fr\'{e}chet Distance maintains stable and consistent scaling across different scenarios. 
As shown in Fig.~\ref{fig:FPD Performance}, JS-divergence is neither consistent nor effective across different disturbance events, while Kernel distance may fail to capture significant changes and is sensitive to kernel choice. Additionally, both methods also exhibit scale variability across disturbances. In contrast, Fr\'{e}chet distance consistently reflects disturbance levels and maintains stable scaling, making it a more reliable evaluation metric.
For these reasons, we select Fr\'{e}chet distance as the evaluation metric and refer to our method as Fr\'{e}chet Power-Scenario Distance (FPD).

% In previous sections, we demonstrated the reason we want to use distribution evaluation metric on extracted feature space over other method metric. And there are three type of method that we can use:(1) KL-divergence (2)Fr\'{e}chet Distance (Wasserstein-2 Distance) and (3)Kernel Distance (Maximum Mean Discrepancy - MMD).

% To decide which metric we want to utilize, we deduct the experiment again on the Gaussian Noise case. According to Fig.\ref{fig:FPD Performance}, KL-divergence is possible to fail because of its asymmetry, and the different comparison directions may have different penalty preferences.

% Kernel Distance has no such issue. However, it may suffer from being less obvious about the difference in the feature space. Besides, the result of Kernel Distance will vary with the choice of kernel type and the parameters of specific kernels. 

% Therefore, considering the metric's stability and consistency, we use Fr\'{e}chet Distance to evaluate the similarity of feature representations.

\subsection{Comparison of FPD with conventional evaluation metrics}
\begin{figure}[htbp]
    \centering
    \includegraphics[width=1\linewidth]{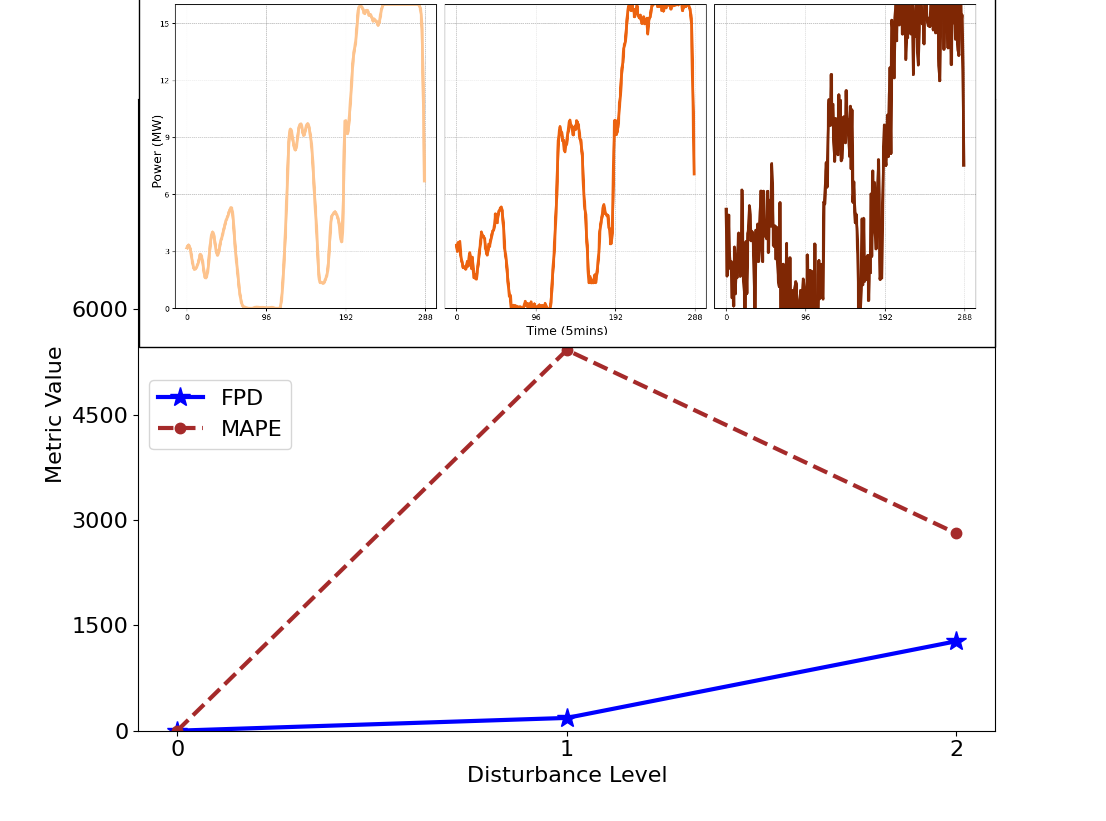}
    \caption{Sample-wise distance MAPE vs. FPD}
    \label{fig:MAPE Vs. FPl}
\end{figure}

\begin{figure}[htbp]
    \centering
    \includegraphics[width=1\linewidth]{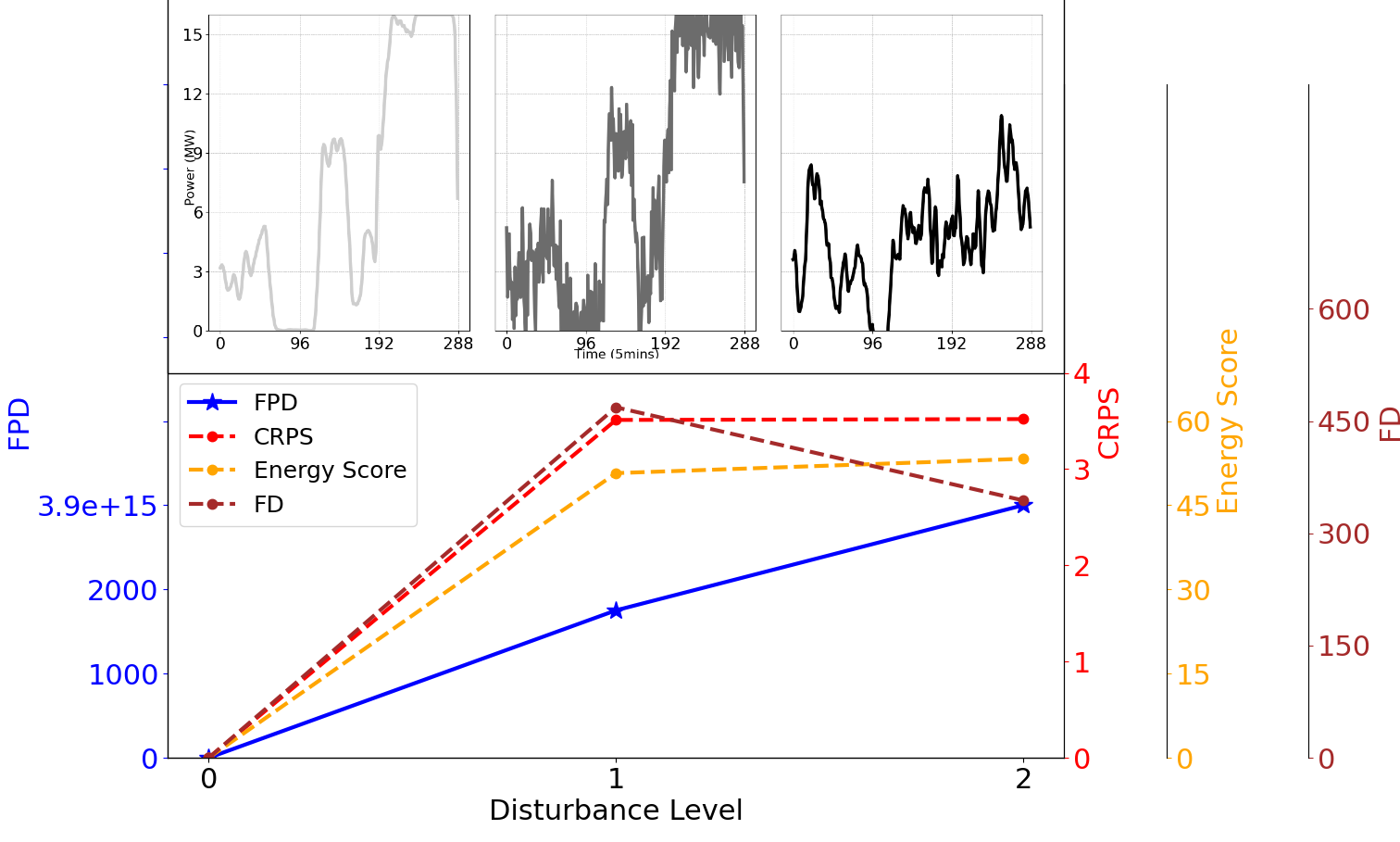}
    \caption{Data-level distribution metrics vs. FPD}
    \label{fig:FD Vs. FPl}
\end{figure}

% \subsection{Compare FPD with traditional metrics}
\subsubsection{Sample-Wise Euclidean Distance (MAPE) vs. Distribution-Wise Distance (FPD)}
Traditional Euclidean-based metrics, such as Mean Absolute Percentage Error (MAPE), typically perform sample-wise comparisons, making the results highly dependent on the selection of sample pairs between the two datasets. This approach can lead to inconsistent and potentially invalid outcomes. To illustrate this, we revisit the Gaussian noise case presented in Fig.~\ref{fig:FPD Performance}. In this example, we compute MAPE after randomly matching sample pairs between the original and noise-perturbed datasets. As shown in Fig.~\ref{fig:MAPE Vs. FPl}, in some instances, higher noise levels result in lower MAPE values, which contradicts the ground truth. In contrast, the proposed FPD metric demonstrates expected behavior, providing a more effective measurement.

\subsubsection{Data-Level Distribution Comparison vs. Feature-Level Distribution Comparison (FPD)}
% As previously noted, the data-level metrics are often ineffective at capturing high-level complex patterns. FPD metric addresses this issue by incorporating feature extraction and measuring the similarity between distributions in the feature space. This highlights the importance of the Feature extraction mechanism we implemented, which enables a better representation of higher-level patterns and structures within the data, allowing FPD to capture more relevant similarities for evaluating the quality of generated data.

% To demonstrate this advantage, we constructed a fabricated dataset by randomly sampling from a distribution with a similar mean and covariance to the original data. Despite these similarities, the fabricated dataset lacks the temporal patterns present in the original dataset, resulting in a sparse relation. As shown in Fig. \ref{fig:FD Vs. FPl}, while FD fails to identify this disparity, FPD accurately reflects the weaker relationship between the fabricated and original datasets.
Although distribution-level scoring rules such as CRPS and the Energy Score, as well as distributional metrics like JS-divergence, MMD, and Wasserstein distance, avoid the sample-wise limitations discussed above, they still struggle to detect changes in generation quality when applied directly to raw data. In particular, such data-level metrics often underemphasize higher-order structure—such as temporal dependencies and cross-feature interactions.

The proposed FPD metric addresses this gap by introducing a feature-extraction stage prior to distributional comparison: data are first mapped into a learned representation, and similarity is then evaluated in this feature space. This feature-based approach highlights higher-level patterns and structures, yielding similarity estimates that more faithfully capture the quality of generated data.

To illustrate, we construct a fabricated dataset by randomly sampling from a distribution with mean and covariance matched to the original data. Despite this alignment in first- and second-order statistics, the fabricated dataset lacks the temporal structure of the original, resulting in weaker dependencies. As shown in Fig.~\ref{fig:FD Vs. FPl}, CRPS and the Energy Score show little separation between the fabricated dataset and a lightly perturbed Gaussian baseline, while the Fr\'{e}chet distance also fails to flag the discrepancy. By contrast, FPD correctly identifies a weaker relationship between the fabricated and original datasets.

\subsection{Robustness of FPD to Labeling Schemes and Sample Budgets for Downstream Tasks}
We ask whether FPD’s indication of data quality reliably predicts downstream utility under realistic variation in labeling and sampling. In practice, given a downstream task, generative models can synthesize effectively unbounded scenario sets whose usefulness depends on both the scenario definition and the sampling strategy. We therefore test whether FPD remains aligned with downstream performance as labeling schemes change and the sampling budget varies. Specifically, we evaluate FPD under alternative wind-power ramp labeling schemes and across multiple sample budgets.

We use the SDWPF dataset \cite{zhou2022sdwpf} with 10-minute resolution and reshape each example as a 1-hour window of six 10-minute values per turbine (\(X \in \mathbb{R}^{N\times6}\)). Let the ramp rate be
\[
r=\frac{P_{50}-P_{00}}{P_{\max}},
\]
where \(P_{00}\) and \(P_{50}\) are the first and last 10-minute values within the hour.

We label SDWPF ramp events by direction and magnitude using two threshold sets inspired by \cite{ela2009wind} and \cite{potter2009potential} as shown in Table~\ref{tab:ramp-rules}.
\begin{table}[t]
\centering
\caption{Ramp-rate category definitions for the two labeling scenarios.}
\label{tab:ramp-rules}
\small
\setlength{\tabcolsep}{3pt}  
\renewcommand{\arraystretch}{1.15}
\begin{tabular}{|l|c|c|}
\hline
\textbf{Category} & \textbf{Scenario 1 (coarser)} & \textbf{Scenario 2 (finer)} \\
\hline
\shortstack{Strong\\down-ramp}    & $r < -0.50$                 & $r < -0.30$ \\ \hline
\shortstack{Moderate\\down-ramp}  & $-0.50 \le r < -0.33$       & $-0.30 \le r < -0.20$ \\ \hline
\shortstack{Mild\\down-ramp}      & $-0.33 \le r < -0.25$       & $-0.20 \le r < -0.10$ \\ \hline
Neutral                           & $-0.25 \le r \le 0.25$      & $-0.10 \le r \le 0.10$ \\ \hline
\shortstack{Mild\\up-ramp}        & $0.25 < r \le 0.33$         & $0.10 < r \le 0.20$ \\ \hline
\shortstack{Moderate\\up-ramp}    & $0.33 < r \le 0.50$         & $0.20 < r \le 0.30$ \\ \hline
\shortstack{Strong\\up-ramp}      & $r > 0.50$                  & $r > 0.30$ \\ \hline
\end{tabular}
\end{table}

We then train two Wasserstein GAN (W-GAN) generators: W-GAN-full on the complete dataset (all categories) and W-GAN-collapse after removing all up-ramping categories to induce mode collapse (the model learns only down-ramp patterns). For each scenario, an identical MLP classifier is trained on real data labeled by that scenario and applied to synthetic samples from each generator at three sizes (0.2k/2k/20k). We report weighted-F1 and macro-F1. Weighted-F1 is the weighted average F1 score, so it reflects performance on prevalent categories and is a proxy for the perceived quality of typical generations. Macro-F1 is the unweighted mean of the F1 score, so it treats all categories equally and is sensitive to diversity. As shown in Table~\ref{tab:wgan-scenarios}, both generators achieve similarly high weighted F1, which indicates high-quality generation on prevalent categories, whereas macro-F1 drops markedly for W-GAN-collapse, revealing it is ignoring up-ramp categories, which is a classic mode collapse.

To compute the FPD score, we first upsample the real and synthetic series to 5-minute resolution, extract hourly features, and then measure the Fr\'{e}chet distance between the generated and real feature distributions. Across both scenarios and all sample sizes, W-GAN-full consistently attains lower (better) FPD than W-GAN-collapse, with values stable as sample size increases. These results demonstrate that FPD aligns with the downstream classification diagnosis and remains robust to the choice of labeling scenario as well as sampling size.

\begin{table*}[h]
\centering
\caption{Classification and FPD (scaled by $10^{-4}$) results under two scenarios}
\label{tab:wgan-scenarios}
\small
\setlength{\tabcolsep}{3.5pt}
\renewcommand{\arraystretch}{1.15}
\begin{tabular}{l *{12}{c}}
\toprule
& \multicolumn{6}{c}{\textbf{Scenario 1}} & \multicolumn{6}{c}{\textbf{Scenario 2}} \\
\cmidrule(lr){2-7}\cmidrule(lr){8-13}
& \multicolumn{3}{c}{\textbf{W-GAN-full}} & \multicolumn{3}{c}{\textbf{W-GAN-collapse}}
& \multicolumn{3}{c}{\textbf{W-GAN-full}} & \multicolumn{3}{c}{\textbf{W-GAN-collapse}} \\
\cmidrule(lr){2-4}\cmidrule(lr){5-7}\cmidrule(lr){8-10}\cmidrule(lr){11-13}
& \shortstack{Weighted\\F1} & \shortstack{Macro\\F1} & FPD
& \shortstack{Weighted\\F1} & \shortstack{Macro\\F1} & FPD
& \shortstack{Weighted\\F1} & \shortstack{Macro\\F1} & FPD
& \shortstack{Weighted\\F1} & \shortstack{Macro\\F1} & FPD \\
\midrule
Sample\_0.2k   & 0.994 & 0.955 & 454.4 & 0.987 & 0.640 & 903.5 & 0.995 & 0.936 & 454.4 & 0.989 & 0.679 & 794.3 \\
Sample\_2k  & 0.998 & 0.984 & 549.1 & 0.980 & 0.602 & 923.3 & 0.994 & 0.918 & 549.1 & 0.989 & 0.673 & 731.7 \\
Sample\_20k & 0.997 & 0.980 & 550.7 & 0.984 & 0.631 & 885.8 & 0.994 & 0.946 & 550.7 & 0.990 & 0.676 & 800.5 \\
\bottomrule
\end{tabular}
\end{table*}

\subsection{Cross-model comparison with FPD}
In the previous sections, all evaluations were conducted directly on the artificially generated datasets. This approach enables precise control over generation quality and model performance assessment. However, FPD can also be used to evaluate model generation performance on the same task, as demonstrated in the following two scenarios.

\begin{enumerate}
    \item \textbf{Case 1: EV Charging Profile Data Generation} 
% \subsubsection{Scenario 1: EV Charging Profile Data Generation}

% Accurately evaluating the quality of synthetic data in real-world settings  presents a significant challenge, as highlighted by \cite{li2024diffcharge}, for which FPD provides a solution. 
In \cite{li2024diffcharge}, the authors present DiffCharge, a diffusion-based generative model that combines an LSTM backbone with multi-head self-attention to conditionally generate EV charging power profiles. They show that DiffCharge better captures short-/long-range temporal dependencies and cross-time interactions than GAN baselines for both battery-level (single-EV sessions) and station-level (aggregate daily load) scenarios. Building on this line of work—and because station-level evaluations were not reported in the underlying study—we train a Wasserstein GAN (W-GAN) baseline on station-level data, compare it against DiffCharge, and assess both using our FPD metric. We also report Kernel distance (RBF) and CRPS. As shown in Fig.~\ref{fig:FPD_EV}, DiffCharge better preserves general temporal structure, whereas W-GAN produces unrealistic, long-lasting peaks. Consistently, FPD yields larger distances for W-GAN outputs—aligning with Kernel distance and CRPS—while offering even more apparent discrimination. Hyperparameter choices for DiffCharge and W-GAN appear in SI Sec.S2; Table S5 and Table S7, respectively.
% Tables~\ref{tab:appB_diffcharge} and~\ref{tab:appB_wgan_merged}, respectively.

% Probability density function (PDF) comparisons and discriminative scores derived from a trained classifier were used to evaluate battery-level data generation performance between DiffCharge and baseline models such as Time-GAN. However, evaluations for station-level data generation were not given. We conduct experiments to compare DiffCharge's performance with a W-GAN model on the station level; see Fig. \ref{fig:FPD_EV}. We find that DiffCharge achieves a lower FPD score. Visually, we also observe that 
% DiffCharge can capture the general trend better, while W-GAN generates less realistic peaky signals.
% % achieves a lower FPD score on the station level as well, signifying its superior ability to capture charging trends accurately. It generates a synthetic signal that is smoother and less noisy compared to W-GAN. Besides, the data generated by W-GAN shows an unrealistic peak duration and magnitude.
% Therefore, our results indeed support the conclusion that the DiffCharge model can more accurately approximate the distribution of EV charging profiles at both the station-level and battery-level.

% \subsubsection{Scenario 2: Wind Power Data Generation}
 \item \textbf{Case 2: Wind Power Data Generation}  
 
To further demonstrate FPD’s cross-model comparison ability, we replicate the study of \cite{yuan2022conditional}, which proposed a StyleGAN that integrates conditions on meteorological variables (e.g., temperature, humidity, weather codes) and employs a sequence encoder to infer day-ahead forecast patterns, yielding more realistic generation than GAN baselines such as the conditional W-GAN of \cite{chen2018model}. In our replication, we randomly sample day-ahead meteorology paired with the corresponding day-ahead forecast as conditioning inputs and generate 1-day wind-power trajectories (96 15-minute intervals) under a mixed weather scenario. As a baseline, we train a conditional W-GAN with the same conditioning and sampling protocol to generate the same horizon. As shown in Fig.~\ref{fig:elia}, samples from StyleGAN track the overall trend and variability of the real data more closely, whereas W-GAN—despite being meteorology-aware—fails to capture the realistic structure in this case. This ranking is correctly reflected by our proposed FPD metric and by CRPS, while a kernel distance computed directly on the raw data fails to distinguish the models. These results underscore the importance of the feature extraction design in FPD. Hyperparameter choices for StyleGAN and W-GAN appear in SI Sec.S2; Table S6 and Table S7, respectively.
% Tables~\ref{tab:appB_stylegan} and~\ref{tab:appB_wgan_merged}, respectively.

% They showed that StyleGAN is superior to WGAN proposed in \cite{chen2018model} based on multiple uncertainty quantification and statistical metrics, such as RMSE. In our experiment, the differences between generations from StyleGAN and WGAN were challenging to distinguish through visual inspection. StyleGAN's results seemed to capture the overall trend more accurately. As shown in Fig. \ref{fig:elia}, FPD confirms StyleGAN aligns better with real data, supporting the findings in \cite{yuan2022conditional}.
% we can spot that StyleGAN result includes more comprehensive scenarios but is overly smooth and misses some finer details. While WGAN result intend to repeat the similar scenario for multiple times.
% \textcolor{red}{The last sentence does not make much sense for me. The sentence before it is also not complete: it is only about the signals, but says nothing about whether the FPD score indicates it.}
% we can spot that both results are good generation because of the relatively small FPD value. The generation difference is difficult to distinguish by visual inspection. \cite{yuan2022conditional} concludes that StyleGAN is superior to WGAN based on multiple uncertainty quantification and statistical metrics, such as RMSE, while FPD reaches the same conclusion with a unified score.
\begin{figure}[htbp]
    \centering
    \includegraphics[width=1\linewidth]{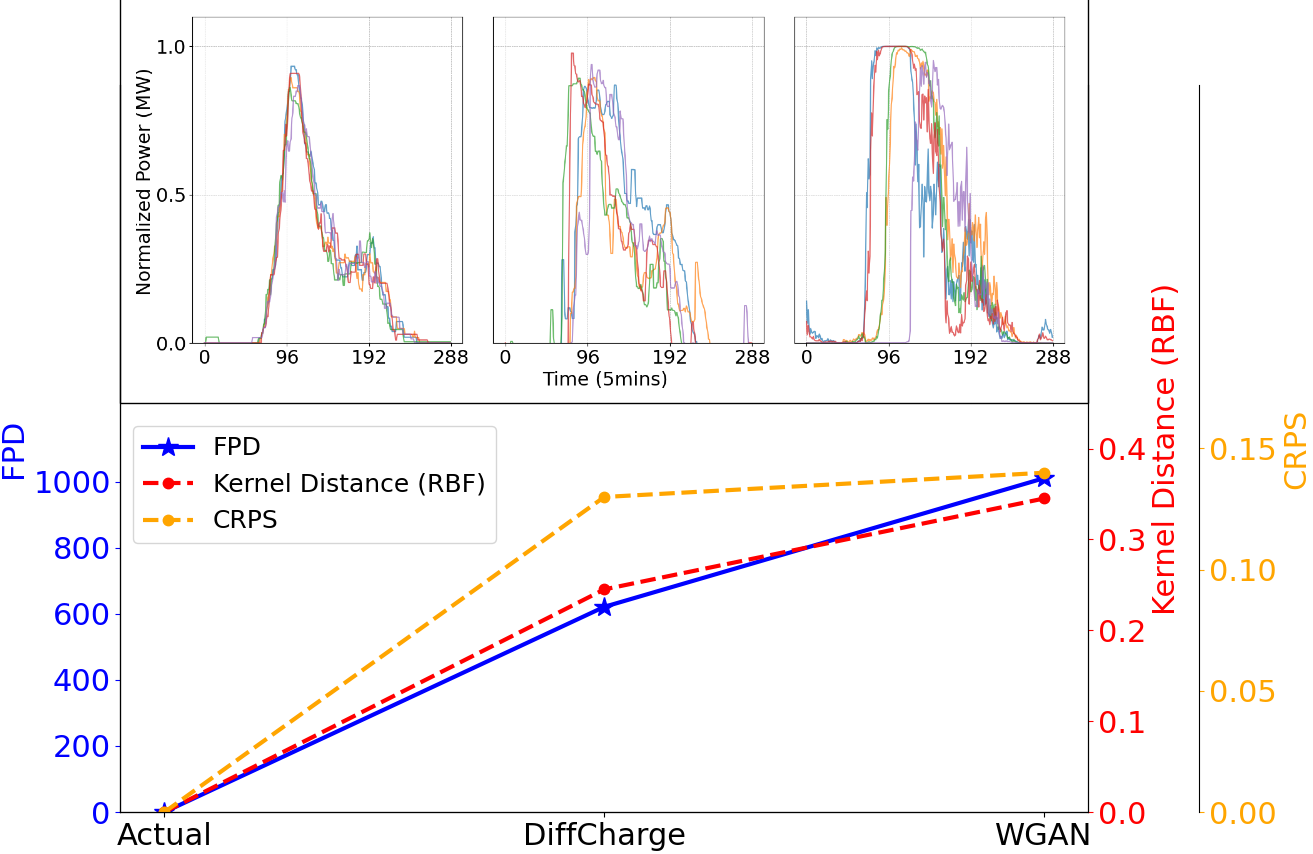}
    \caption{\textbf{Case 1} – Comparison of DiffCharge and W-GAN in generating EV charging profiles at the station level.}
    \label{fig:FPD_EV}
\end{figure}

\begin{figure}[htbp]
    \centering
    \includegraphics[width=1\linewidth]{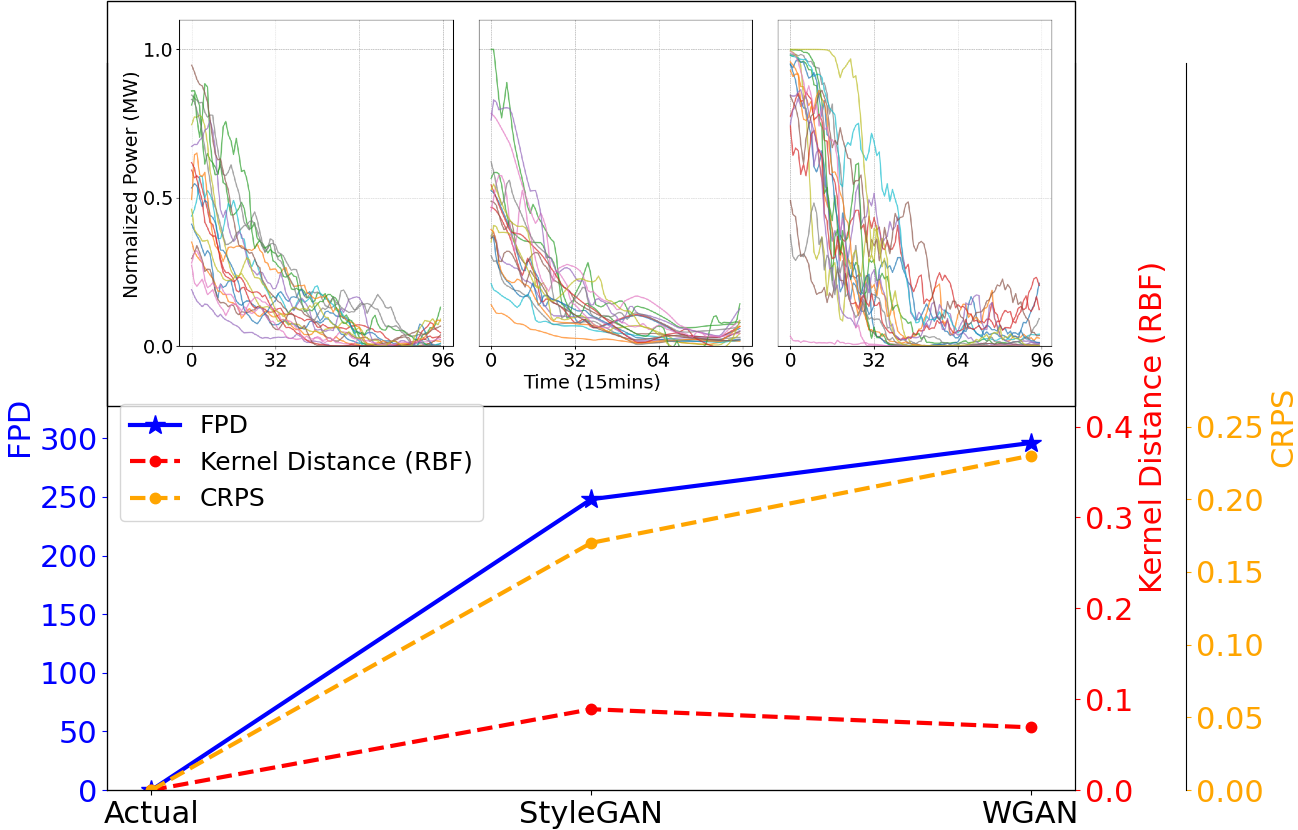}
    \caption{\textbf{Case 2} – Comparison of W-GAN and StyleGAN in generating wind power data.}
    \label{fig:elia}
\end{figure}
\end{enumerate}

\section{Conclusion and Future Work}
\label{section:conclusion}
This paper introduces the Fr\'{e}chet Power-Scenario Distance (FPD)—a feature-based, distributional metric tailored to smart grid data for evaluating synthetic datasets. FPD encodes both generic temporal structure and domain-specific properties before computing distances in the learned feature space. This design yields consistent, effective judgments across disturbances, time resolutions, and scenario definitions, and better aligns with downstream utility, where raw-data metrics and generic encoders often fall short.

% FPD addresses traditional limitations by using a hierarchical feature extractor network to capture high-level, multi-resolution temporal structure, followed by a Fr\'{e}chet-distance comparison in the resulting feature space. This method enables FPD to provide an accurate, robust, and consistent evaluation across diverse smart grid applications.

Comprehensive studies demonstrate FPD's effectiveness in distinguishing data quality across various disturbances, time resolutions, and sub-task scenarios. These results underscore FPD's potential as a standardized cross-model approach for validating synthetic datasets, offering support for both enhancing model quality and promoting wider adoption within the smart grid domain.

There are several directions for future work. 1) Currently, the dataset we are using is still limited due to accessibility constraints. Thus, building a more comprehensive dataset tailored to a broader range of tasks within the smart grid domain could improve the performance of FPD; 2) While the current model structure can handle multi-resolution and multi-duration data, it is not yet sufficiently general to accommodate data of any resolution or duration. Therefore, there is potential for developing a more generalized structure that can address this challenge and further enhance the versatility of the metric.
% This paper introduced the Fr\'{e}chet Power-Scenario Distance (FPD), a novel evaluation metric tailored to assess generative models in power systems across multiple time scales. FPD overcomes the limitations of traditional metrics by integrating both spatial and temporal features, enabling a standardized and task-agnostic evaluation framework. Our method leverages hierarchical feature extractors at various time scales, making it highly applicable to diverse tasks such as renewable energy scenario generation and synthetic PMU data profiling.

% Through comprehensive case studies, we demonstrated FPD’s ability to effectively differentiate all kinds potential problem raised in generation model, offering a more robust and generalizable assessment than existing methods. This advancement paves the way for more reliable evaluations of generative models in power systems, with future work aimed at expanding its applications across additional domains.

% \section*{Acknowledgments}
% This should be a simple paragraph before the References to thank those individuals and institutions who have supported your work on this article.
\newpage
\bibliographystyle{unsrt}
\bibliography{main}

\clearpage
% \section*{Supplementary Information}
% \addcontentsline{toc}{section}{Supplementary Information} % optional
\includepdf[pages=-,pagecommand={}]{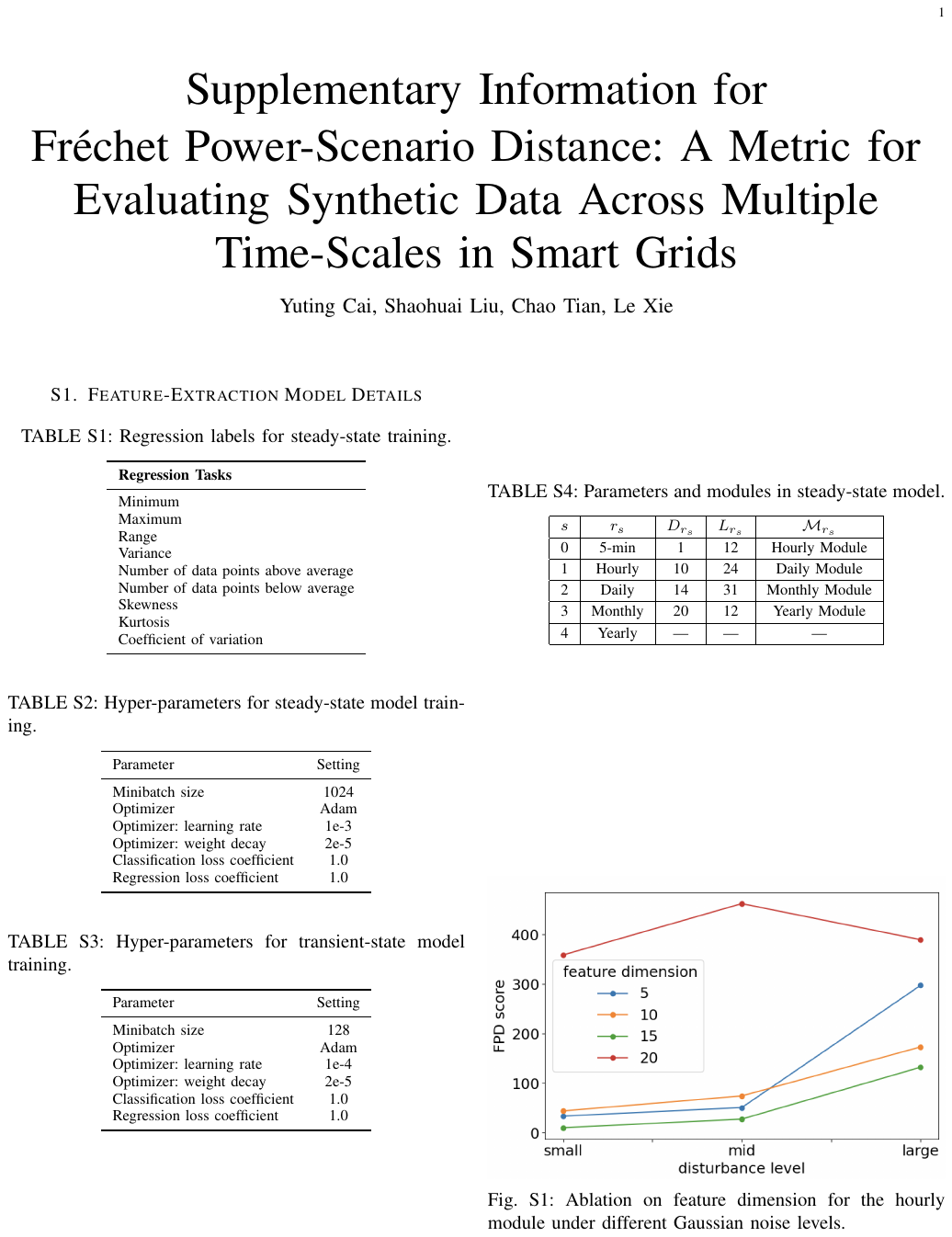}

\vfill

\end{document}